\documentclass{article}

\usepackage[preprint]{tmlr}
\usepackage[utf8]{inputenc}
\usepackage{amsmath,amssymb,amsfonts}
\usepackage{mathtools} 
\usepackage{graphicx}
\usepackage{booktabs}
\usepackage[hidelinks]{hyperref} 
\usepackage{url}
\usepackage{microtype}
\usepackage{xcolor}
\usepackage{subcaption}
\usepackage{array}

\newcolumntype{L}[1]{>{\raggedright\arraybackslash}p{#1}}

\title{Weight Decay Regimes in Grokking Transformers:\\Cheap Online Diagnostics}

\author{\name Lucky Verma \email luckyv1@umbc.edu \\
        \addr Independent Researcher}

\date{}

\begin{document}

\maketitle

\begin{abstract}
Grokking transformers trained on modular arithmetic exhibit sharp transitions between memorization, generalization, and collapse regimes. We show that weight decay acts as a scalar empirical control parameter for these regimes, and we introduce two cheap online diagnostics, mean pairwise attention-head cosine similarity and entropy standard deviation, that track the underlying training dynamics in these decoder-only modular-arithmetic settings from attention activations alone and complement loss-landscape diagnostics at lower compute cost. Across eleven experimental conditions and three model scales (0.82M to 85M parameters), the weight-decay axis separates memorization ($\lambda<\lambda_c$, near-zero grokking), developmental grokking ($\lambda\geq\lambda_c$, reaching $\sim$100\% grok rate by $\lambda\sim 0.1$ with time-to-grok decreasing $1090\to83$ epochs over $\lambda\in[0.1, 2.0]$), and collapse ($\lambda=10$, identical attention patterns). A near-transition logistic fit localizes the memorization-to-developmental boundary at $\lambda_c=0.0158$ (95\% CI $[0.0109, 0.0200]$, $N{=}210$); a power-law fit to time-to-grok gives an empirical exponent $\nu=0.757$ (CI $[0.725,0.799]$). Tested reference exponents $\nu=1/2$ and 3D Ising $\nu\approx0.63$ lie outside the empirical CI under our four-bin grid; we report $\nu$ as empirical and defer universality-class identification to denser finite-size-scaling data collapse~\citep{bi2026grokking}. A horizon-matched multi-task replication ($n{=}280$, four modular operations, two scales) preserves the WD-control pattern beyond addition; a paired attention-head re-initialization experiment changes Phase-2 amplitude at canonical post-transition $\lambda=0.05$ (Cohen's $d=-1.190$, $n{=}10$ paired, $p_t=4.5{\times}10^{-3}$), while matched weight-norm clipping does not, isolating the effect to head-pattern structure rather than weight magnitude. Per-head dimension $d/H$ modulates differentiation amplitude saturating-monotonically. Three cross-architecture scope probes (4L MLP $h{=}512$, 4L LSTM $h{=}512$, and 4L Mamba $d{=}128$; each canonical $n{=}70$) replicate the WD-controlled transition in non-attention architectures, with per-architecture $\lambda_c$ spanning roughly an order of magnitude (MLP $\lambda_c{=}0.0511$ $[0.0495, 0.0591]$; LSTM $\lambda_c{=}0.0365$ $[0.0299, 0.0473]$; Mamba $\lambda_c{=}0.0144$ $[0.0106, 0.0159]$, whose CI overlaps the transformer canonical CI rather than sitting above it). Claims are scoped to modular arithmetic in small transformer attention models; architecture-wide, language-model, and universality-class claims are out of scope.
\end{abstract}

\section{Introduction}
\label{sec:intro}

When transformers are trained on modular arithmetic, they exhibit \emph{grokking}: a sudden transition from memorization to generalization long after training accuracy saturates~\citep{power2022grokking}.
Mechanistic analysis reveals that this transition corresponds to the formation of internal circuits (Fourier-based representations) that emerge gradually but manifest abruptly~\citep{nanda2023progress}.
Similarly, the formation of induction heads during transformer training constitutes a ``phase change'' visible from two-layer models to 70B+ parameter systems~\citep{olsson2022context}.

Across individual training phenomena (grokking~\citep{power2022grokking,nanda2023progress,kumar2024grokking}, emergent abilities~\citep{wei2022emergent}, lottery tickets~\citep{frankle2019lottery}, neural collapse~\citep{papyan2020prevalence}), recent learning-mechanics perspectives frame staged learning, empirical laws, limiting models, and cheap measurable diagnostics as central objects for a future theory of deep learning~\citep{simon2026scientific}.
The statistical mechanics community has connected training to phase transitions~\citep{bahri2020statistical,saxe2014exact,ziyin2023phase,zunkovic2024statistical}, and recent work applies oscillator-style synchronization models to neural activations~\citep{miyato2025akorn}.
However, existing grokking studies rarely provide a cheap online activation-space diagnostic and regime map for weight-decay-controlled attention dynamics.

We make four contributions:

\begin{enumerate}
    \item \textbf{Quantitative weight-decay critical threshold with bootstrap CIs and formal well-formedness checks.} A dense weight-decay sweep plus a sparse three-size scale probe gives a horizon-matched logistic transition estimate $\lambda_c = 0.0158$ (95\% CI $[0.0109, 0.0200]$, $N{=}210$) for the canonical $\mathrm{mod}_+$ dense WD cohort, and a power-law exponent $\nu = 0.757$ (CI $[0.725, 0.799]$) on time-to-grok above $\lambda_c$. The two-axis $(\lambda, N)$ map exhibits three qualitatively distinct regimes (memorization, developmental, collapse) with both ``too little'' and ``too much'' failure modes, complementing finite-size-scaling and weight-geometry work that also treats weight decay as a phase-relevant variable but does not pin down a numerical threshold. Four diagnostic well-formedness identities (A1, B1, C1, E1) are machine-checked in Lean~4; they certify bounds and algebraic identities of the diagnostics, not the empirical regime claims.

    \item \textbf{Cheap online order parameters for attention-head coordination.} We define two scalar quantities computable at every training step: mean pairwise cosine similarity $\bar s(t)$ and entropy standard deviation $\sigma_H(t)$. These track attention-head coordination through training, complement the refined local learning coefficient~\citep{wang2024rllc} while requiring only forward-pass attention weights, and we benchmark head-to-head on identical checkpoints.

    \item \textbf{Two sequential phases plus seed-dependent late-stage retention failure.} Along the canonical training trajectory we document: Phase 1 (attention-head coordination, near grokking) where heads converge and $\bar s$ rises $0.93 \to 0.995$; Phase 2 (differentiation, post-grokking) where heads diverge $\bar s \to 0.88$ while accuracy holds. At twenty thousand training epochs the canonical seed-42 trajectory extends into a five-stage pattern exhibiting late-stage accuracy collapse reminiscent of anti-grokking; the 20K-epoch long-horizon retention cohort (E8, $n{=}20$) shows this is a seed-dependent fragility, not a universal cycle (retention rates $5/5$, $4/5$, $3/5$, $4/5$ at $\lambda\in\{0.1, 0.5, 1.0, 2.0\}$). The collapse trajectory is qualitatively similar to the late-stage cycle reported by~\citet{prakash2026antigrokking}, though our spectral evidence (\S\ref{sec:results:perm-sym}) differs from theirs in timing.

    \item \textbf{Per-head dimension as amplitude modulator.} At fixed model dimension, varying the number of heads isolates per-head dimension $d/H$ as an empirically dominant variable for differentiation amplitude: peak $\sigma_H$ increases saturating-monotonically with $d/H$ (monotone over $d/H\in\{2,4,8,16\}$, plateauing at $d/H{\geq}16$ where bin means at $d/H{=}16$ and $d/H{=}32$ have overlapping 95\% bin-CIs), reaching the same order of magnitude as random-label null controls at $d/H \approx 2$ while remaining statistically distinguishable (permutation-test $p{=}0.009$, Cohen's $d{=}1.11$). We therefore label this an empirical architectural threshold in this setting and defer causal-mechanism claims to follow-up work.
\end{enumerate}

\section{Related Work}
\label{sec:related}

\paragraph{Grokking and training phase transitions.}
\citet{power2022grokking} discovered that small transformers generalize on algorithmic tasks long after memorization, a phenomenon explained mechanistically by \citet{nanda2023progress} as circuit formation followed by cleanup, and theoretically by \citet{varma2023explaining} as competition between memorizing and generalizing circuits driven by weight decay.
\citet{kumar2024grokking} frame grokking as a lazy-to-rich regime transition, while \citet{liu2023omnigrok} extend grokking beyond algorithmic data.
Work from 2025 and 2026 has converged on framing grokking as a quantitative phase transition: \citet{bi2026grokking} apply finite-size scaling with Binder-cumulant crossings and spectral head-tail contrast as order parameter; \citet{wang2026dimensional} and \citet{wang2026dimensionalcrit} analyze effective dimensionality and self-organized criticality via cascade-dimension exponents; \citet{acharya2026variance} connect grokking to variance-limited spectral gating; \citet{khanh2026entropy} propose normalized spectral entropy of the representation covariance as a scalar threshold (crossing $\sim\!0.61$ before generalization); \citet{hennick2026density} study reduced-density-matrix spectra as early warnings; \citet{golwala2026ildr} uses held-out representation-centroid geometry for early detection; \citet{tian2025li2} provide provable three-stage scaling laws in two-layer networks.
\citet{xu2026ridgegrokking} prove delayed generalization in ridge regression under weight decay, while \citet{zhang2026abstraction} give an SLT/algorithmic-complexity abstraction view of grokking; \citet{song2026capacity} relate grokking on modular arithmetic to competing memorisation and generalisation timescales as functions of parameter count, complementary to our empirical WD$\times N$ map under a fixed protocol. These results reinforce the value of cheap order parameters and weight-decay-controlled regimes, but do not supply the WD$\times N$ attention-head regime map studied here.
\citet{lyu2023dichotomy} establish the canonical early/late-phase implicit-bias dichotomy; \citet{musat2025geometry} characterize grokking as norm minimization on the zero-loss manifold driven by weight decay; \citet{manir2026systematic} find empirically that grokking is primarily determined by regularization and optimization rather than architecture.
\citet{prakash2026antigrokking} report a ``previously unreported third phase'' (late-stage generalization collapse) diagnosed via spectral density heavy-tailedness, with follow-up RMT/Correlation-Trap framing in \citet{prakash2026rmtgrok} treating anti-grokking as a long-horizon overfitting phase. We observe a qualitatively similar late-collapse trajectory using attention-similarity diagnostics and characterize its weight-decay dependence, though our spectral evidence (\S\ref{sec:results:perm-sym}) differs from theirs because heavy-tail structure forms during grokking onset rather than during the late cycle.

\paragraph{Concurrent work on grokking geometry.}
\citet{xu2026multitask} studies multi-task grokking on modular arithmetic and identifies weight decay as a \emph{phase parameter} that governs grokking timescale and curvature depth, reporting two qualitative regimes ($\lambda \!\geq\! 0.5$ fast vs.\ $\lambda \!\leq\! 0.3$ slow) with complementary results in~\citet{xu2026lowdim,xu2026earlywarning,xu2026spectraledge,xu2026spectralifecycle,xu2026functionaledge}.
Our work differs in three respects.
First, \citet{xu2026multitask} holds model size $N$ fixed throughout; we sweep both $\lambda$ and $N$ jointly, exposing a horizon-matched transition estimate $\lambda_c$ stable across the small/medium pair tested with overlapping 95\% CIs (fit $\lambda_c = 0.0158$, 95\% CI $[0.0109, 0.0200]$, from our dense WD-sweep cohort of $N{=}210$ runs post Phase A) and a third \emph{collapse} regime at $\lambda \!>\! 5$ absent from Xu's diagram.
Second, Xu's diagnostics operate in weight and update space (PCA trajectory variance, the commutator norm $\|[W_Q,W_K]\|_F$, spectral-edge gradient/decay decomposition, functional-mode spectra) and require full checkpoint access; our order parameters (mean pairwise cosine similarity of attention activations and entropy standard deviation across heads) are activation-space diagnostics computable online in $O(H^2)$ per evaluation step.
Third, Xu does not invoke synchronization or permutation-symmetry-reduction dynamics; the Phase~1 synchronization and Phase~2 head-specialization framing, the five-stage anti-grokking trajectory, and the $d/H$ amplitude modulation are not present in their work.
\citet{yildirim2026geometric} presents a counter-framing in which grokking is bypassable via a uniform-attention architectural ablation without weight decay; we do not claim that weight decay is \emph{necessary} for generalization, only that within the standard attention architecture weight decay behaves as an empirical control parameter with a well-defined regime diagram.
\citet{tang2026topology} report a sharp rise in $H_1$ persistent-homology features at grokking onset on modular arithmetic, an offline post-hoc geometric diagnostic complementary to our online attention-coordination order parameters; the two diagnostic classes target the same regime transitions at different signal locations (representation topology vs head-pattern coordination) and different evaluation cost. \citet{wang2026distspectral} likewise localize grokking transitions via distributional spectral coordinates (Wasserstein/quantile, Hankel DMD residual, effective rank) on modular-addition Transformer trajectories; this is another transition-localization diagnostic at a different signal location (weight/activation spectra) rather than an attention-head activation order parameter.
\citet{ali2026windows} examine WD-placement timing on compositional tasks and report a critical training window phenomenon explicitly absent on modular arithmetic, an orthogonal axis (when-WD) to our amount-based regime mapping (how-much-WD).
\citet{gomezjurado2026delayed} attribute delayed generalization in encoder-decoder Collatz prediction to a representation-access bottleneck rather than feature-acquisition failure, complementary to our weight-decay-controlled attention-coordination signal in decoder-only modular-arithmetic models.
\citet{lyle2025grokking} connect grokking-style feature-learning dynamics to nonstationary continual-learning primacy bias via effective learning rate, framing grokking as a pump for plasticity orthogonal to our weight-decay regime diagram.

\paragraph{Emergent abilities in LLMs.}
\citet{wei2022emergent} documented $\sim$100 abilities appearing sharply with scale, though \citet{schaeffer2023emergent} argued some are metric artifacts.
The quantization model of \citet{michaud2023quantization} hypothesizes discrete skill acquisition, while \citet{nam2024solvable} provide analytically tractable models of ordered emergence.

\paragraph{Attention head specialization.}
\citet{voita2019analyzing} showed that few heads carry most computation, while \citet{michel2019sixteen} demonstrated 70 to 90\% of heads are prunable.
\citet{olsson2022context} documented a ``phase change'' during training where induction heads form abruptly.
\citet{chen2024unveiling} prove that induction-head formation follows gradient flow in the infinite-time limit (continuous convergence; not a claim of discrete loss drops in finite-step SGD).
Most directly related to our Phase~2 finding, \citet{wang2024rllc} introduce the refined local learning coefficient (rLLC) as an SGLD-estimated local-learning-coefficient diagnostic for staged head differentiation during training; we benchmark our cosine-similarity-based order parameter against rLLC on the identical 11-checkpoint canonical 4L8H trajectory (epochs 100 to 20\,000, devinterp 1.0.0 SGLD, 3 chains $\times$ 200 draws per ckpt) and find Pearson correlation $r{=}0.46$ between $\mathrm{LLC}(t)$ and $\sigma_H(t)$ (the $n{=}11$ ckpt benchmark gives a wide Fisher-$z$ 95\% CI $[-0.20, 0.82]$ that is consistent with anything from no correlation to a strong one), indicating the two diagnostics are \emph{complementary rather than equivalent}: $\sigma_H$ tracks attention-entropy dispersion across heads (a representation-space signal), while rLLC tracks local-loss-landscape geometry, and the two correlate only moderately across the five phases at the ckpt-count tested. We therefore do not claim $\sigma_H$ replaces rLLC; instead it provides a cheap, online-computable complement that captures a related but distinct aspect of Phase-2 dynamics.
\citet{sagitova2026specialization} provide a theoretical account of softmax-head specialization as a high-dimensional staged transition in a single-location model.

\paragraph{Statistical mechanics of deep learning.}
\citet{bahri2020statistical} connected loss landscapes to spin glasses and dynamical phase transitions.
\citet{saxe2014exact} derived exact learning dynamics for deep linear networks showing plateau-transition structure.
\citet{saxe2019semantic} used these dynamics to model semantic development with stage-like transitions matching developmental psychology data, though restricted to linear networks and toy datasets.
\citet{poole2016exponential} showed networks at the ``edge of chaos'' achieve exponential expressivity.

\paragraph{Synchronization physics in neural networks.}
\citet{miyato2025akorn} replaced threshold neurons with Kuramoto oscillators (AKOrN), demonstrating improved robustness and reasoning through phase-based synchronization.
\citet{nguyen2024kuramoto} applied the Kuramoto model to prevent over-smoothing in GNNs.
\citet{hays2026selective} replaces standard self-attention with a closed-form Kuramoto steady-state operator that turns each token into a learnable-frequency oscillator and reads attention weights from phase-locking strength.
These approaches all apply Kuramoto to \emph{representations} at inference time; we apply it as a qualitative analogy to \emph{training dynamics} over time and explicitly disclaim a quantitative Kuramoto-model fit (\S\ref{sec:methods:kuramoto}).

\paragraph{Developmental landscapes and weight decay mechanism.}
\citet{boukacem2024waddington} demonstrated Waddington-like sequential bifurcations in generalized Hopfield networks on MNIST; we extend this staged-dynamics framing to transformer training dynamics while noting that our observed post-grokking transition is an empirical head-specialization event, distinct from the saddle-node-of-saddles bifurcation of~\citet{boukacem2024waddington}.
The mechanistic role of weight decay has been clarified by recent work: \citet{andriushchenko2023wd} provide a modern account of why weight decay is needed in deep learning; \citet{galanti2025wdrank} show that SGD with weight decay secretly minimizes effective rank; \citet{singh2026inductive} demonstrate that architectural choices such as layer-normalization placement strongly modulate grokking dynamics; \citet{wang2024adamwwd} derive scaling rules for AdamW weight decay across model and dataset size by treating learned weights as an exponential moving average of recent updates, complementing our trajectory-based amplification fit. \citet{khanh2026delay} derive a quantitative scaling law for the grokking delay $T_{\text{grok}} - T_{\text{mem}} = \Theta((1/\gamma_{\text{eff}})\log(\|\theta_{\text{mem}}\|^2/\|\theta_{\text{post}}\|^2))$ via Lyapunov contraction, with $\gamma_{\text{eff}} \ge \eta\lambda$ for AdamW; our \S\ref{sec:appendix-derivation} provides the empirical calibration of the AdamW amplification ($\kappa$) for the canonical 4L8H mod-add cohort, plus the inverted formulation as a $\lambda_c$ threshold under a fixed training horizon. \citet{zhang2025glassrelaxation} frame grokking as a glass-physics relaxation analogically (memorisation as rapid cooling into a glassy state, generalisation as slow relaxation), which is a different abstraction from the mechanistic AdamW-update relaxation argument we deploy in \S\ref{sec:appendix-derivation}.

\paragraph{Biological cardiac synchronization (terminology origin only).}
The staged-coordination vocabulary used in this paper was originally motivated by cardiac synchronization literature: \citet{jia2023bioelectrical} showed the first vertebrate heartbeat is a saddle-node-on-invariant-circle (SNIC) bifurcation; \citet{nitsan2016mechanical} demonstrated mechanical extracellular-matrix-mediated synchronization; \citet{chiou2016heartbeat} showed embryonic hearts coordinate mechanically before electrical infrastructure matures. We make no biological claim; \S\ref{sec:discussion} explains why the empirical scaling exponent rules out the SNIC analogy.

\section{Methods}
\label{sec:methods}

\subsection{Task and Model}
We train small decoder-only transformers on modular arithmetic following \citet{power2022grokking}.
For $\mathrm{mod}_+$, $\mathrm{mod}_\times$, and $\mathrm{mod}_-$, inputs are pairs $(a,b) \in \{0,\dots,p-1\}^2$ with $p=97$ (so $p^2 = 9409$ total examples), half used for training and half held out under a seed-controlled permutation. For $\mathrm{mod}_\div$, pairs with $b=0$ are excluded before the same split, because division by zero is undefined modulo prime $p$.
The canonical architecture is a 4-layer, 8-head transformer with $d_{\text{model}}{=}128$, FFN width 512, dropout 0, pre-LayerNorm.
We additionally sweep layers $\in\{2,4,6,12\}$, heads $\in\{4,8,16,32,64\}$, and $d_{\text{model}}\in\{32,64,128,256,512,768\}$ for the finite-size-scaling and scale-axis experiments, yielding parameter counts from 0.82\,M (small, $4{\times}8{\times}128$, $d_{\text{ff}}{=}512$) through 19\,M (medium, $6{\times}8{\times}512$, $d_{\text{ff}}{=}2048$) to 85\,M (large, $12{\times}12{\times}768$, $d_{\text{ff}}{=}3072$).
All runs use AdamW at $\mathrm{lr}{=}10^{-3}$, batch 512, cross-entropy loss. The canonical $(p, L, H, d, \mathrm{lr}, \mathrm{batch}) = (97, 4, 8, 128, 10^{-3}, 512)$ choice follows the Power 2022 modular-arithmetic grokking baseline used by \citet{nanda2023progress,kumar2024grokking,bi2026grokking}, ensuring direct comparability with the 2026 grokking-wave literature; we sweep $L, H, d_{\mathrm{model}}$ for finite-size-scaling but hold $p$, $\mathrm{lr}$, and batch fixed. Sensitivity to alternative regularizers (dropout, label smoothing) and optimizers (SGD, Lion, Sophia) is out of scope and deferred.
Seeds are drawn from a 10-seed base set
$\mathcal{S}_0 = \{7, 11, 31, 42, 73, 97, 123, 199, 401, 977\}$
at $n{=}10$ per cell. A 20-seed Phase-A extension
$\mathcal{S}_1 = \{2,\allowbreak 3,\allowbreak 5,\allowbreak 17,\allowbreak 23,\allowbreak 29,\allowbreak 53,\allowbreak 67,\allowbreak 89,\allowbreak 103,\allowbreak 113,\allowbreak 137,\allowbreak 149,\allowbreak 167,\allowbreak 181,\allowbreak 197,\allowbreak 211,\allowbreak 229,\allowbreak 241,\allowbreak 263\}$
adds replication at $\lambda \in \{0.01, 0.1, 1.0\}$ for the $n{=}30$ near-critical and canonical cells.

\subsection{Order Parameters}
\label{sec:methods:order-params}
Let $A_{lh}(t) \in \mathbb{R}^{B\times T\times T}$ be the attention-weight tensor of head $h$ in layer $l$ at training step $t$, where $B$ is the evaluation batch size and $T{=}3$ is the token sequence length after appending the output-query token.
We define two scalar order parameters, evaluated every 10 training steps:
\begin{align}
\bar s(t) &\coloneqq \mathbb{E}_{l}\!\left[\tfrac{2}{H(H-1)}\sum_{i<j}\cos\!\big(\mathrm{vec}(A_{li}),\mathrm{vec}(A_{lj})\big)\right] \quad &&\text{(pairwise head cosine)}\\
\sigma_H(t) &\coloneqq \mathbb{E}_{l}\!\left[\mathrm{Std}_h\!\left(H[A_{lh}]\right)\right] \quad &&\text{(entropy std over heads)}
\end{align}
Here $H[\cdot]$ is Shannon entropy. Both diagnostics use only the attention weights already produced by the forward pass; $\bar s$ and $\sigma_H$ cost $\mathcal{O}(L H^2 B T^2)$ and $\mathcal{O}(L H B T^2)$ respectively per evaluation step, compared with checkpoint-level SGLD estimation for the rLLC diagnostic of \citet{wang2024rllc} (dominated by repeated model evaluations and sampling chains rather than head-count). Since $T=3$ is fixed in these modular-arithmetic runs, the diagnostic overhead is dominated by batch size and head count rather than model parameters. On the canonical 4L8H model, evaluating $\bar s$ and $\sigma_H$ adds approximately $3\%$ wall-clock overhead amortized over the every-10-step cadence used here. Two complementary controls (Kuramoto coherence $r_\phi$ and similarity-matrix spectral gap $\lambda_g$) are tracked in supplementary trace JSONs but not used for the regime map or causal contrasts; their definitions and selection rationale are in Appendix~\ref{sec:appendix-extra-diagnostics}.

\subsection{Synchronization Analogy and Tanh Fit}
\label{sec:methods:kuramoto}
As a loose synchronization analogy, one can treat each head as an oscillator whose phase is the principal attention-pattern direction.
Empirically we fit the pre-grokking rise of $\bar s(t)$ to the overdamped tanh form $\bar s(t) = s_0 + A\tanh((t-t_c)/\tau)$, which matches the shape of a uniform-coupling mean-field synchronization model with time-independent drive.
Of $n{=}50$ canonical runs (post Phase A; expanded from $n{=}22$ pre-Phase-A), the unconstrained tanh fit achieves $R^2{>}0.9$ in 10 runs and $R^2{>}0.7$ in 33 runs (median $R^2{=}0.843$). However, $16/50$ of those unconstrained fits achieve high apparent $R^2$ via $\tanh$ saturating to a constant ($s_0,A$ diverging with opposite signs over the fit window), which is a fit-form degeneracy on a near-flat Phase~1 plateau rather than a successful synchronization fit. Restricting to physically interpretable parameters ($s_0\in[0,1.5]$, $A\in[0,1]$, $|t_c|,\tau<5000$) yields $34/50$ valid fits with median $R^2{=}0.63$ and only $4/34$ at $R^2{>}0.9$. This supports only a qualitative synchronization analogy for Phase~1; it is not quantitative validation of a Kuramoto model, and explicit coupling-constant extraction from the AdamW update is deferred.

\subsection{Finite-Size Scaling}
\label{sec:methods:fss}
Following finite-size-scaling analyses of grokking~\citep{bi2026grokking, wang2026dimensional}, we test whether time-to-grok near $\lambda_c$ follows $t_{\text{grok}} \propto (\lambda-\lambda_c)^{-\nu}$.
We estimate $\lambda_c$ by fitting a logistic to $P(\mathrm{grok})$ vs $\log \lambda$ across $N{=}210$ small-scale runs (logistic cohort post Phase A; expanded from $N{=}150$ pre-Phase-A via $n{=}30$ replication at $\lambda \in \{0.01, 0.1, 1.0\}$), then fit $\nu$ by linear regression of $\log t_{\text{grok}}$ on $\log(\lambda-\lambda_c)$ restricted to grokking runs.
The scale axis $N \in \{0.82, 19, 85\}\,$M is sampled at $\lambda \in \{0.01, 0.1, 1.0\}$ with $n{=}10$ seeds per cell. This supports the empirical $\lambda \times N$ phase map but remains insufficient for full finite-size data collapse because the weight-decay grid has only three bins (instrumented but underpowered; denser grids and $n{\geq}20$ per cell targeted for future work).
Confidence intervals are reported from nonparametric bootstraps (1{,}500 resamples for the logistic $\lambda_c$ and power-law $\nu$ fits) and from leave-one-seed-out jackknife; the residual-bootstrap sensitivity check on $\nu$ uses 5{,}000 resamples.

\section{Results}
\label{sec:results}

\subsection{Two-axis regime diagram}
\label{sec:results:phase-diagram}
\begin{figure}[t]
\centering
\includegraphics[width=\linewidth]{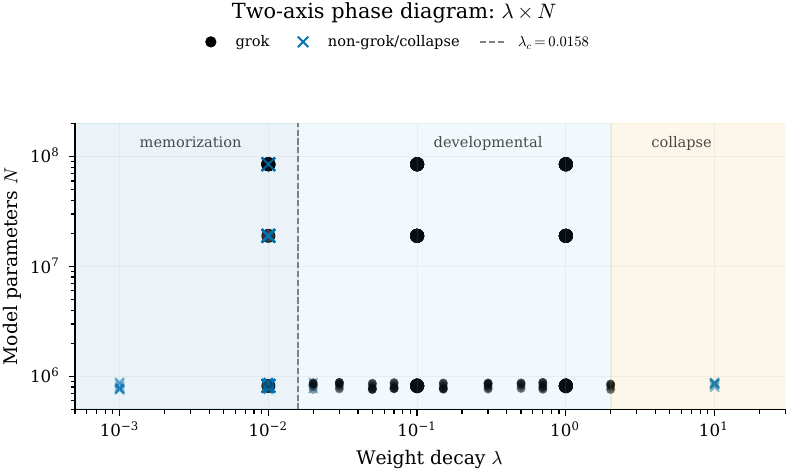}
\caption{Two-axis empirical regime diagram across $n{=}300$ runs post Phase A (210 WD-sweep logistic cohort + 90 three-scale runs). Filled black circles: grokking runs. Blue crosses: non-grokking (memorization or collapse). Shaded regions: regimes along the $\lambda$ axis. Dashed vertical line: $\lambda_c{=}0.0158$ logistic fit.}
\label{fig:phase-diagram}
\end{figure}

Figure~\ref{fig:phase-diagram} maps the empirical $\lambda \times N$ regime diagram from $N{=}210$ WD-sweep runs (post Phase A) and $n{=}90$ three-scale sweep runs (E5; 3 sizes $\times$ 3 WDs $\times$ 10 seeds).
Along the WD axis at small $N$: $\lambda{<}0.0158$ yields near-zero grokking; $\lambda \in [0.1, 2.0]$ gives roughly 90 to 100\% grokking with monotone time-to-grok decrease $1090 \to 83$ epochs; $\lambda = 10$ collapses all heads to identical patterns ($\bar s = 1.000$ exact).
Along the $N$ axis at $\lambda = 1.0$: small (0.82\,M) groks reliably; large (85\,M) collapses into a null state ($\bar s = 1.000$, $\sigma_H = 0.000$) by epoch 3000.
The available scale grid therefore supports scale dependence under the stated training horizons, but not a monotone $\lambda_c(N)$ law; the completed horizon-matched small/medium follow-up (E7) and multi-task pooled refit (E9) are used only to rule out a clean small-to-medium monotone-threshold claim, not to identify a new scale law.

\subsection{Two-phase dynamics and five-phase long-horizon cycle}
\label{sec:results:two-phase}
\begin{figure}[t]
\centering
\includegraphics[width=\linewidth]{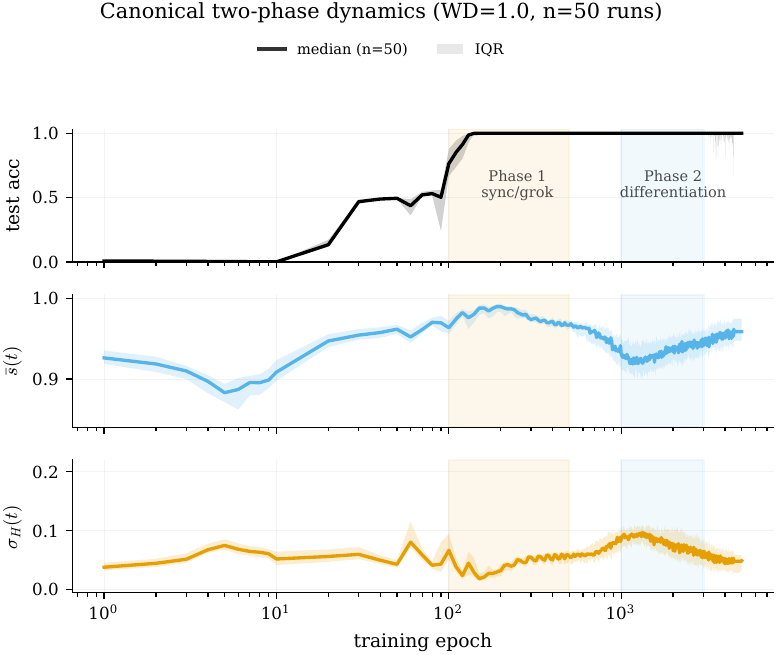}
\caption{Canonical two-phase dynamics at $\lambda{=}1.0$ over $n{=}50$ replicated 4L8H runs. Solid curves show cohort medians; bands show interquartile ranges. Phase~1 (yellow shading): synchronization + grokking. Phase~2 (blue shading): differentiation + test-accuracy plateau.}
\label{fig:two-phase}
\end{figure}

\begin{figure}[t]
\centering
\includegraphics[width=\linewidth]{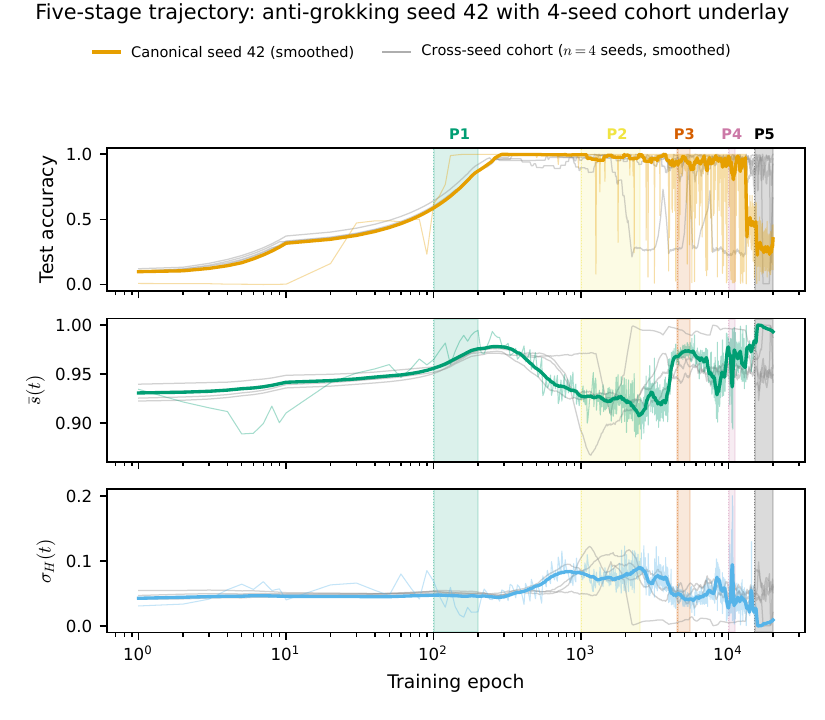}
\caption{Long-horizon canonical seed-$42$ trajectory with a $4$-seed matched underlay ($\{7, 11, 31, 123\}$) at 4L8H mod-add $\lambda{=}1.0$ $20\,000$-epoch configuration. Bold colored traces (canonical seed $42$): raw per-epoch values at low alpha plus moving-average-smoothed overlay. Gray traces: smoothed cross-seed trajectories underlaid to disclose cohort variability. P1--P5 bands are canonical seed-$42$ landmark windows, not per-seed fitted boundaries. P1: attention-head coordination. P2: first differentiation. P3: re-sync. P4: second differentiation (canonical raw $\sigma_H$ peak $0.201$ at epoch $\sim$10\,500; smoothed peak $\sim$0.10). P5: collapse with canonical test-accuracy decay to $0.46$. The five-stage pattern is most pronounced for the canonical seed; the cross-seed underlay shows that 1--2 of the four additional seeds exhibit terminal late-cycle accuracy collapse and the remaining seeds recover or retain generalization by the full horizon, consistent with the seed-dependent fragility documented in the E8 retention cohort.}
\label{fig:five-phase}
\end{figure}

Figure~\ref{fig:two-phase} shows the replicated canonical cohort ($n{=}50$) exhibiting sharp Phase~1 synchronization (median $\bar s$ rises during epochs 100 to 200) followed by Phase~2 differentiation (IQR-widening in $\bar s$ and $\sigma_H$ during epochs 1000 to 3000) while test accuracy remains near plateau.
At 20\,000 epochs (Figure~\ref{fig:five-phase}) the canonical seed-42 trajectory elongates into five annotated stages terminating in anti-grokking collapse ($\bar s \to 0.99$, test accuracy $\to 0.46$); a $4$-seed cohort underlay (seeds $\{7, 11, 31, 123\}$ at matched 4L8H $\lambda{=}1.0$ $20\,000$-epoch configuration via the cross-seed checkpoint cohort) plotted alongside the canonical detail discloses cross-seed variability directly. The horizon-matched long-horizon retention cohort (E8; $n{=}5$ seeds at each of $\lambda \in \{0.1, 0.5, 1.0, 2.0\}$, $20$ runs total at 20\,000 epochs) confirms the five-stage pattern is a \emph{seed-dependent fragility} rather than a universal cycle: retention rates are $5/5$ at $\lambda{=}0.1$, $4/5$ at $\lambda{=}0.5$, $3/5$ at $\lambda{=}1.0$, and $4/5$ at $\lambda{=}2.0$, so collapse concentrates at higher WD but a non-trivial fraction of seeds retains generalization across the full horizon at every WD tested. The five-stage canonical figure shows the strongest cycle pronouncement, occurring in roughly one third of canonical-WD seeds; the cross-seed overlay is the direct visual confirmation. The pattern is qualitatively consistent with the third-phase phenomenology of \citet{prakash2026antigrokking}.

\subsection{Per-head dimension as amplitude modulator}
\begin{figure}[t]
\centering
\includegraphics[width=0.6\linewidth]{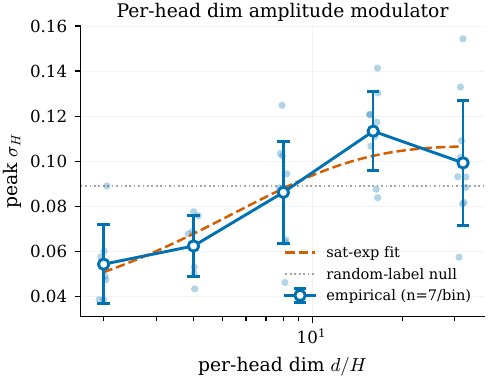}
\caption{Peak $\sigma_H$ vs per-head dimension $d/H$. Saturating-exponential AIC-preferred. Dotted line: random-label null-control scale reference, not an equivalence claim.}
\label{fig:dh}
\end{figure}

At fixed total model dimension $d$, varying head count $H$ scans per-head dimension $d/H$.
Figure~\ref{fig:dh} shows peak $\sigma_H$ following a saturating-exponential profile $\sigma_H^{\max} = c+a(1-e^{-b(d/H)})$ across $n{=}44$ small-scale runs (Akaike information criterion, AIC, preferred over linear, log, and power-law alternatives), approaching the random-label-null scale $0.087\pm0.025$ at $d/H \approx 2$.

\subsection{Scaling exponent and universality-class limits}
\label{sec:results:exponents}
\begin{figure}[t]
\centering
\includegraphics[width=\linewidth]{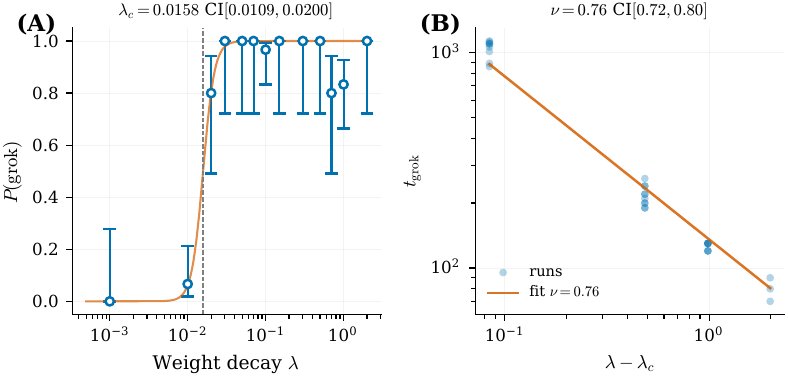}
\caption{(A) Logistic $P(\mathrm{grok})$ vs weight decay across $N{=}210$ runs post Phase A. Points are WD bins with Wilson 95\% CIs, the orange curve is the logistic fit, and the dashed vertical line marks $\lambda_c{=}0.0158$ (95\% CI $[0.0109, 0.0200]$). (B) Power-law divergence of $t_{\text{grok}}$ above $\lambda_c$; $\nu{=}0.757$, CI $[0.725, 0.799]$. Tested reference exponents such as $\nu{=}1/2$ and 3D Ising $\nu{\approx}0.63$ are outside CI under the four-bin grid; we do not identify a universality class.}
\label{fig:wdc-nu}
\end{figure}

Figure~\ref{fig:wdc-nu}A shows the logistic $P(\mathrm{grok})$ fit locating $\lambda_c = 0.0158$ (95\% CI $[0.0109, 0.0200]$) across $N{=}210$ WD-axis runs (logistic cohort, 13 WD bins with $n{=}10$ to $30$ seeds per bin, Phase A replication at $\lambda \in \{0.01, 0.1, 1.0\}$).
Figure~\ref{fig:wdc-nu}B shows the power-law $t_{\text{grok}} \propto (\lambda-\lambda_c)^{-\nu}$ fit with $\nu = 0.757$ (95\% CI $[0.725, 0.799]$, $n{=}140$ grok-positive runs in the $N{=}210$ logistic cohort); a jackknife on the multi-task-extended grok-positive cohort ($n{=}148$) gives bias-corrected $\nu{=}0.761$ ($[0.723, 0.799]$) with residual-bootstrap CI $[0.738, 0.786]$. These $\nu$ intervals are conditional on the fitted point estimate $\lambda_c=0.0158$ and do not propagate uncertainty in $\lambda_c$ itself; a fully joint $(\lambda_c, \nu)$ bootstrap is deferred to the denser-grid finite-size-scaling work cited next.
The measured exponent does not match tested reference exponents such as $\nu{=}1/2$ or three-dimensional Ising $\nu{\approx}0.63$ (both outside CI under our four-bin grid). We report the value as an \emph{empirical exponent} and explicitly defer universality-class identification to future finite-size-scaling data-collapse work with denser weight-decay grids and larger per-cell replication.

\subsection{Permutation-symmetry reduction: direct test on canonical checkpoints}
\label{sec:results:perm-sym}
Head indices are exchangeable at initialization (permutation group $S_H$ acts transitively on heads); post-Phase-2, head-pattern covariance becomes lower-participation and permutation-distinguishable. We instrument this directly on the 11 saved canonical-trajectory checkpoints (4L8H, $d{=}128$, $\lambda{=}1.0$, 20\,000 epochs, seed $42$) using the affine-normalized participation ratio of the head-covariance spectrum:
\begin{equation}
R_l(t) \coloneqq \frac{\big(\sum_{k=1}^H a_{lk}(t)\big)^2}{\sum_{k=1}^H a_{lk}(t)^2},
\qquad
\mathrm{PR}_{\text{norm}}(t) \coloneqq
\mathbb{E}_l\!\left[\frac{R_l(t)-1}{H-1}\right],
\end{equation}
where $a_{lk}(t)$ are the nonnegative eigenvalues of the layer-$l$ head-pattern covariance matrix used in the direct test. $\mathrm{PR}_{\text{norm}}=1$ when all spectral directions participate equally, and $\mathrm{PR}_{\text{norm}}=0$ under rank-1 concentration.

\begin{figure}[t]
\centering
\includegraphics[width=0.75\linewidth]{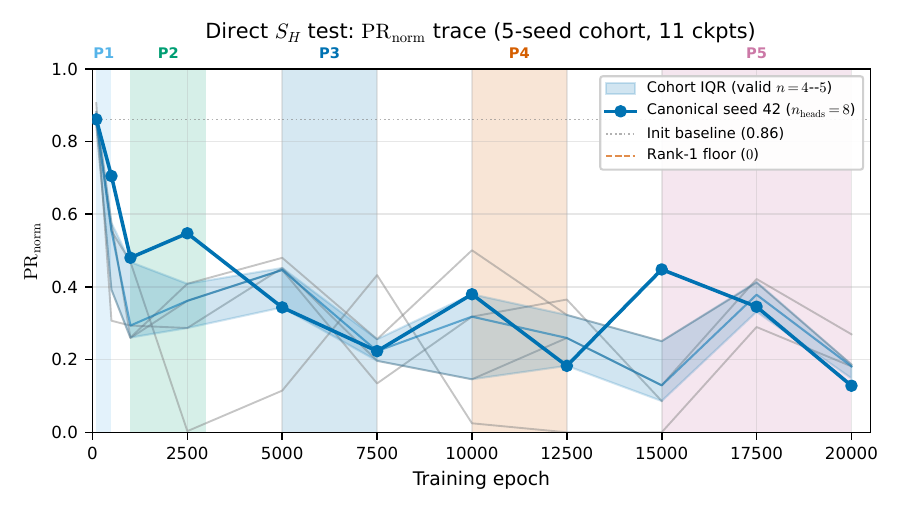}
\caption{Permutation-symmetry test via affine-normalized $\mathrm{PR}_{\mathrm{norm}}$ on 11 canonical-trajectory checkpoints across a $5$-seed cohort (canonical seed $42$ + cross-seed cohort $\{7, 11, 31, 123\}$, all at 4L8H $\lambda{=}1.0$ $20\,000$-epoch matched configuration). Blue circles + bold line: canonical seed $42$ focus trace. Gray thin lines: cross-seed individual traces. Blue shaded band: cohort interquartile range over finite values at each checkpoint (valid $n{=}4$ at epoch $17\,500$ and $n{=}5$ otherwise). Gray dotted: random-init baseline $0.86$. Vermilion dashed: rank-1 floor $0$. The canonical median-across-layers trace falls to $0.71$ at Phase~1 onset, oscillates between $0.18$ and $0.55$ across Phases 2 to 4, and collapses to $0.13$ at Phase~5; the cross-seed cohort confirms the qualitative trajectory across all five seeds with seed-level scatter at the late-Phase nadir. Cross-seed C1-identity validation appears in Appendix~\ref{sec:appendix-c1-validation}; Table~\ref{tab:c1-validation} reports the layer-averaged values.}
\label{fig:pr-norm}
\end{figure}

The canonical seed-42 trace (Figure~\ref{fig:pr-norm}, 11 ckpts, $\mathrm{PR}_{\text{norm}}$ reported to 2 s.f.): initialization $0.86 \to$ Phase~1 attention-head coordination $0.71$ at epochs 100 to 500 (grokking onset) $\to$ Phases~2 to 4 differentiation oscillation (epochs 1000 to 12\,500) $\to$ Phase~5 collapse $0.13$ at epoch 20\,000 (low-participation head-covariance structure, well below the random-initialization baseline). A $5$-seed cohort (canonical seed $42$ + cross-seed $\{7, 11, 31, 123\}$) plotted in Figure~\ref{fig:pr-norm} confirms the same qualitative trajectory across all five seeds with seed-level scatter at the late-Phase nadir; the IQR band uses finite values at each checkpoint (valid $n{=}4$ at epoch $17\,500$ and $n{=}5$ otherwise). This replaces the proxy $M_{\text{perm}}$ of the pre-checkpoint manuscript with a direct $S_H$-breaking diagnostic. Appendix~\ref{sec:appendix-c1-validation} verifies the raw participation-ratio/CV identity and the released affine normalization on 183 layer-epoch rows from the same five seeds, with maximum absolute PR error $1.73{\times}10^{-6}$.

\begin{figure}[t]
\centering
\includegraphics[width=0.75\linewidth]{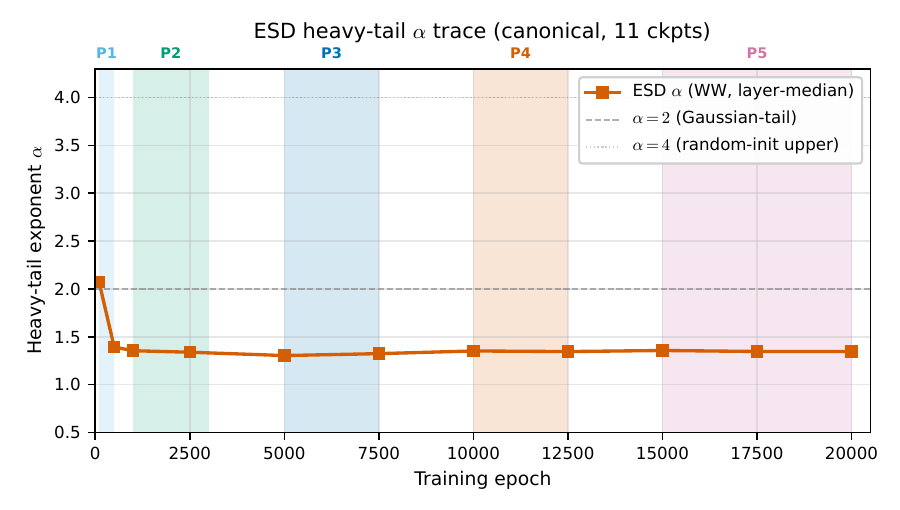}
\caption{ESD heavy-tail exponent $\alpha$ (Weightwatcher, layer-median) across the 11 canonical-trajectory checkpoints (seed $42$ only). Cross-seed Weightwatcher at the matched 11-checkpoint grid is deferred; a coarser 3-seed supplementary trace provides a qualitative onset check. $\alpha$ drops from $2.07$ at random init to $1.39$ at Phase~1 grokking onset and remains in the heavy-tail regime ($\alpha < 2$) through Phase~5. The third-phase signature from \citet{prakash2026antigrokking} co-occurs with grokking onset rather than developing during late-stage collapse, complicating its use as a direct collapse-onset detector.}
\label{fig:esd-alpha}
\end{figure}

We additionally run spectral empirical-density (Weightwatcher) analysis on the same 11 ckpts, tracking the heavy-tail exponent $\alpha$ of the weight-matrix spectral density, following the Weightwatcher heavy-tail framework~\citep{martin2021weightwatcher} as applied to anti-grokking by~\citet{prakash2026antigrokking}; the trace is shown in Figure~\ref{fig:esd-alpha}. $\alpha$ falls from $2.07$ at initialization to $1.39$ by epoch 500 (Phase~1 grokking onset) and remains $\lesssim 1.5$ through Phase~5. Heavy-tail structure therefore \emph{forms during grokking onset, not during late-stage collapse}, complicating direct identification of the ``third phase'' signature from \citet{prakash2026antigrokking} with our trajectory: the two phenomena co-occur but $\alpha$ is not itself monotone with generalization collapse. We report this as a partial parallel rather than a reproduction.

\subsection{Causal intervention: head re-initialization vs weight clipping}
\label{sec:causal-intervention}
\paragraph{Pre-specified hypotheses.}
The causal-intervention study (E12) was evaluated against three planned hypotheses (set before outcome inspection, not externally registered): H1, head re-initialization reduces peak $\sigma_H$ relative to the paired control; H2, the intervention does not prevent grokking; and H3, the weight-clipping control distinguishes head-pattern disruption from a matched weight-norm perturbation. The pooled paired contrasts are confirmatory for these hypotheses. WD-stratified contrasts, including the sub-critical $\lambda{=}0.015$ cell, are exploratory scope checks.

The order-parameter results in Sections~\ref{sec:results:two-phase} through~\ref{sec:results:perm-sym} are correlational: peak $\sigma_H$, $\bar s$, and $\mathrm{PR}_\mathrm{norm}$ all evolve together during Phase~2. To probe causal structure we run a paired-design intervention experiment ($n{=}60$ runs, 3 groups $\times$ 10 seeds $\times$ 2 intervention $\lambda$ values): group~A is unintervened control, group~B re-initializes 2/8 attention heads at the per-seed peak-$\sigma_H$ epoch (typically iterations 1{,}000 to 2{,}000), group~C clips $\Vert W \Vert_2$ to the per-seed median at the same epoch. All other hyperparameters are matched.

All three groups grok at $100\%$ rate (10/10 in groups A and B at both $\lambda$ values; 9/9 in group C at $\lambda{=}0.05$, where one weight-clip run was excluded because the post-clip optimizer state produced a non-finite gradient on the next backward pass; the exclusion was logged before downstream analysis and the decision was applied identically across cells). The intervention does not prevent generalization. Phase~2 amplitude does respond causally: paired across seeds and $\lambda$ ($n{=}20$), the peak~$\sigma_H$ difference $B-A$ has mean $-0.038 \pm 0.043$ (paired $t$, $p_t{=}9.3{\times}10^{-4}$, Wilcoxon $p{=}2.5{\times}10^{-3}$, Cohen's $d{=}-0.876$).\footnote{Paired $t$ tests assume approximately normal within-pair differences and can be affected by outliers; Wilcoxon signed-rank tests are the non-parametric paired check and give the same headline decision here. Cohen's $d$ is computed on paired differences, with $0.2/0.5/0.8$ as small/medium/large reference magnitudes. Residual-bootstrap CIs for the $\nu$ sensitivity fit use 5\,000 resamples; the jackknife reports a leave-one-out bias-corrected $\nu$.} The peak per-head dimension differential $\bar r_{\mathrm{diff}}$ shows a smaller paired effect ($\Delta{=}-0.014$, $p_t{=}0.020$, $d{=}-0.569$), final test accuracy and $\min(\bar s)$ are unchanged, and grokking epoch shifts by $377{\pm}1\,282$ iterations ($p_t{=}0.20$, statistically null at this $n$).

Stratified by $\lambda$ ($n{=}10$ per cell), the head-re-initialization effect on peak~$\sigma_H$ is concentrated at the canonical post-transition $\lambda{=}0.05$ ($\Delta{=}-0.055 \pm 0.046$, $p_t{=}4.5{\times}10^{-3}$, $d{=}-1.190$) and is non-significant at sub-critical $\lambda{=}0.015$ ($\Delta{=}-0.021 \pm 0.034$, $p_t{=}0.086$, $d{=}-0.609$). Group C (weight clipping) yields no significant peak~$\sigma_H$ change relative to group A at either $\lambda$ (paired $d{=}-0.406$, $p_t{=}0.094$ pooled), localising the effect to head-pattern structure rather than weight norm. The C-vs-B contrast at canonical $\lambda{=}0.05$ is significant ($\Delta_{\mathrm{peak}~\sigma_H}{=}+0.048 \pm 0.030$, $p_t{=}1.4{\times}10^{-3}$, $d{=}+1.586$): replacing head structure suppresses Phase~2 amplitude in a way that scaling weight norm does not reproduce.
Across the 15-test paired-outcome family (three pooled pairwise contrasts $B{-}A$, $C{-}A$, $C{-}B$ across five outcomes: peak $\sigma_H$, minimum $\bar s$, peak $\bar r_{\mathrm{diff}}$, grokking-onset epoch, and final test accuracy), the strict Bonferroni cutoff is $\alpha/15=0.0033$; the pooled $B-A$ peak-$\sigma_H$ result remains below this cutoff, while the planned WD-stratified follow-up retains the canonical $\lambda{=}0.05$ contrast under Holm-Bonferroni and the sub-critical $\lambda{=}0.015$ contrast does not survive correction.

\begin{figure}[t]
\centering
\includegraphics[width=0.82\linewidth]{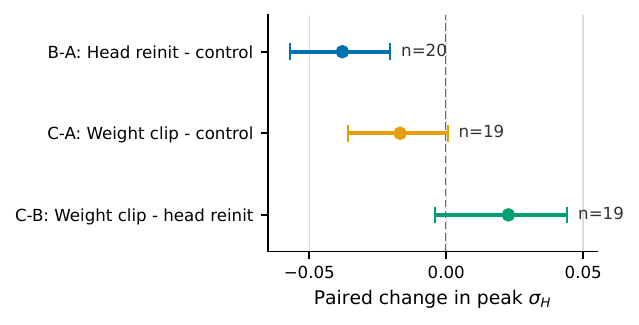}
\caption{Causal-intervention paired forest plot for peak $\sigma_H$, pooled across $\lambda\in\{0.015, 0.05\}$. Points show mean paired change and bars show 95\% CIs: $B-A=-0.038$ ($[-0.057,-0.020]$, $n{=}20$), $C-A=-0.017$ ($[-0.036,0.0005]$, $n{=}19$), and $C-B=+0.023$ ($[-0.004,0.044]$, $n{=}19$). WD-stratified breakdown in \S\ref{sec:causal-intervention} body: at canonical post-transition $\lambda{=}0.05$, $B-A=-0.055$ ($p_t=4.5{\times}10^{-3}$, $d=-1.190$) and $C-B=+0.048$ ($p_t=1.4{\times}10^{-3}$, $d=+1.586$) are both highly significant.}
\label{fig:intervention-forest}
\end{figure}

We read this as causal evidence (forest plot in Figure~\ref{fig:intervention-forest}) that Phase~2 amplitude is sensitive to head-pattern structure rather than weight magnitude, at canonical post-transition $\lambda$ in the studied 4L8H~$d{=}128$ transformer on $\mathrm{mod}_+$ ($p{=}97$). We do not claim the head-perturbation effect generalizes across architectures, tasks, or WD values outside $\{0.015, 0.05\}$, nor that intervention rescues any retention metric.

\subsection{Controls and predictions}
\label{sec:results:controls}
A random-label null control ($n{=}15$ extended random-label control runs, superseding the smaller E8 pilot) yields peak $\sigma_H{=}0.087\pm0.025$, the same order of magnitude as the $d/H{=}2$ low-amplitude regime; the add-vs-random amplitude contrast is significant but small (permutation-test $p{=}0.009$, Cohen's $d{=}1.11$, $n_a{=}12$ vs $n_b{=}15$), so the null control is a meaningful lower bound on amplitude rather than statistically equivalent to $d/H{=}2$.

\begin{figure}[t]
\centering
\includegraphics[width=\linewidth]{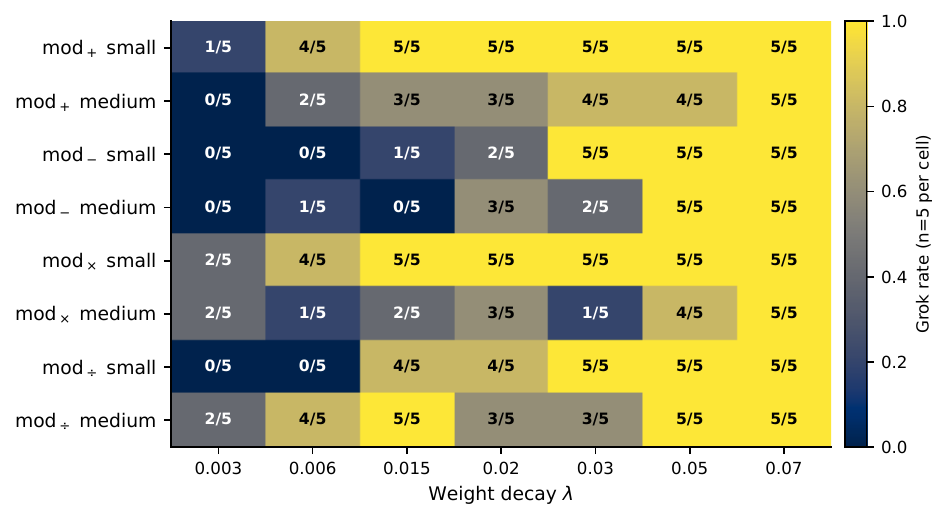}
\caption{Multi-task grok-rate heatmap (4 modular operations $\times$ 2 model scales $\times$ 7 weight decays, $n{=}5$ seeds per cell, $N{=}280$ total). Each cell shows $\mathrm{groked}/n$; color encodes the same fraction on a cividis sequential scale. Rows are mostly monotone left-to-right: weight decay is the dominant axis. Pooled task ordering $\mathrm{mod}_+\,0.73$, $\mathrm{mod}_\div\,0.71$, $\mathrm{mod}_\times\,0.70$, $\mathrm{mod}_-\,0.49$ reproduces the sub-hardness ranking of the original $n{=}28$ task-control cohort.}
\label{fig:multitask-grok}
\end{figure}

\paragraph{Multi-task replication at $n{=}280$.}
The earlier $n{=}28$ task-control cohort (E3) is now superseded by a horizon-matched multi-task sweep (E9) (Figure~\ref{fig:multitask-grok}) across all four operations $\mathrm{mod}_+$, $\mathrm{mod}_-$, $\mathrm{mod}_\times$, $\mathrm{mod}_\div$ at two scales (4L8H $d{=}128$ and 6L8H $d{=}512$) and seven near-critical $\lambda$ values with $n{=}5$ seeds each, yielding $n{=}280$ runs at 10\,K epochs. Per-WD pooled grok rate (across 4 ops $\times$ 2 scales) increases monotonically from $7/40$ ($0.175$) at $\lambda{=}0.003$ to $40/40$ ($1.00$) at $\lambda{=}0.07$, reproducing the canonical WD-control structure. Per-task grok rate (pooled WDs and scales) is $\mathrm{mod}_+\,0.73$, $\mathrm{mod}_\div\,0.71$, $\mathrm{mod}_\times\,0.70$, $\mathrm{mod}_-\,0.49$, recovering the $\mathrm{mod}_-$ slow-grok ordering of the original E3 cohort at much larger $n$. Pooled-4-ops logistic refit yields $\lambda_c$ small $0.0077$ (95\% CI $[0.0056, 0.0106]$, $n{=}102/140$ grok) and $\lambda_c$ medium $0.0128$ ($[0.0076, 0.0194]$, $n{=}82/140$). The small and medium CIs overlap, consistent with our horizon-matched verdict that no monotone $\lambda_c(N)$ law is supported within the small/medium pair. The E9 pooled-4-ops small CI does not overlap with the WD-axis canonical CI $\lambda_c{=}0.0158$ ($[0.0109, 0.0200]$); this is a protocol difference, not a contradiction. The canonical headline is fit on the $\mathrm{mod}_+$ dense WD-sweep cohort with 13 WD bins at the canonical training horizon; E9 is the horizon-matched pooled four-operation cohort with 7 WD bins at $10{,}000$ epochs. The two estimates agree on the qualitative monotone WD-control structure and on the order of magnitude of $\lambda_c$; they differ in numerical point estimate because they pool different tasks and use different WD-bin densities. We treat $\lambda_c{=}0.0158$ as the canonical mod-add reference and the E9 pooled values as task-pool consistency probes, not as competing point estimates of a single underlying scalar. Per-task medium-scale spread is suggestive but individually under-powered at $n{=}5$ per cell, with mutually overlapping 95\% CIs ($\mathrm{mod}_\div\,0.003$, $\mathrm{mod}_+\,0.012$, $\mathrm{mod}_\times\,0.019$, $\mathrm{mod}_-\,0.025$); the pooled monotone WD-control pattern rather than the task-ordering carries the inference. The eight per-task logistic fits (4 tasks $\times$ 2 scales) are treated as a single multiple-comparison family; the active inference is the pooled monotone WD-control pattern rather than an individually significant task-ordering test.

\subsection{Order parameters as correlational discriminators of long-horizon retention}
\label{sec:results:retention-discrim}
We tested whether the cheap online order parameters defined in \S\ref{sec:methods:order-params} predict long-horizon outcome (stable retention vs.\ never-grok vs.\ collapse) using a 50-run held-out cohort (held-out retention test) at WD values not seen in the training set ($\lambda\in\{0.025, 0.04\}$) with new seeds. Train cohort: $998$ runs from the Phase-A pool; features per run at epoch~1\,K: $\sigma_H$, $\bar s$, weight norm, $\mathrm{ent}_\mathrm{std}$, peak $\bar r_{\mathrm{diff}}$, $d/H$, scale, WD, test accuracy. Two classifiers (logistic regression and random forest) were grouped-CV-trained at the (scale, WD) level.

Random forest holdout area under the ROC curve (AUC) is $0.799$ with Brier score $0.098$ on the canonical scikit-learn 1.5.x training environment; across recent scikit-learn versions (1.4.x to 1.6.x) on identical inputs the holdout AUC ranges from $0.79$ to $0.81$ and Brier from $0.098$ to $0.102$ due to default-parameter changes in \texttt{RandomForestClassifier}, with both AUC extrema below the pre-specified $0.85$ predictor gate. Logistic regression holdout AUC is $0.678$. Train-time AUC was $0.834$, so holdout generalization gap is $-0.035$ (roughly $4\%$ relative drop). The verdict from our pre-specified acceptance gate (AUC$\ge0.85$ for predictor status, set before outcome inspection and not externally registered) is \textbf{correlational only}: the order parameters at epoch~1\,K carry retention signal but do not reach the predictor threshold on out-of-distribution WDs. Top-five RF feature importances are WD ($0.247$), weight norm ($0.220$), $\bar s$ ($0.145$), $\mathrm{ent}_\mathrm{std}$ ($0.130$), test accuracy ($0.123$); the order-parameter features ($\bar s$, $\mathrm{ent}_\mathrm{std}$) together account for $0.275$ of feature importance, comparable to weight-norm magnitude alone.

\subsection{Cross-architecture scope probes: 4L MLP, 4L LSTM, and 4L Mamba grids}
\label{sec:results:cross-arch}
\begin{figure}[t]
\centering
\includegraphics[width=0.78\linewidth]{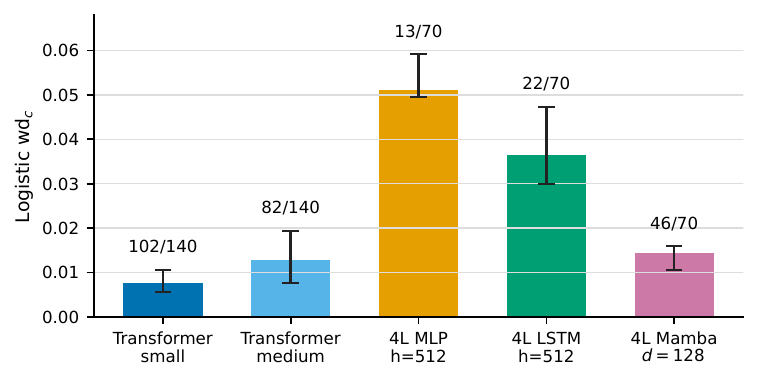}
\caption{Cross-architecture transition comparison from logistic grok-rate fits. Bars show $\lambda_c$ and 95\% bootstrap CIs for transformer small ($0.0077$, CI $[0.0056,0.0106]$, $102/140$ grok), transformer medium ($0.0128$, CI $[0.0076,0.0194]$, $82/140$), 4L MLP $h{=}512$ ($0.0511$, CI $[0.0495,0.0591]$, $13/70$), 4L LSTM $h{=}512$ ($0.0365$, CI $[0.0299,0.0473]$, $22/70$), and 4L Mamba $d{=}128$ ($0.0144$, CI $[0.0106,0.0159]$, $46/70$). The $\lambda_c$ values span roughly an order of magnitude across the five architecture configurations; reported as empirical observation, no mechanism claimed.}
\label{fig:cross-arch}
\end{figure}

To bound architecture-specificity we ran a horizon-matched cross-architecture sweep (E10) (Figure~\ref{fig:cross-arch}) on a 4-layer MLP (hidden dimension $512$, no attention) for $\mathrm{mod}_+$, $p{=}97$, with 7 WD values $\times$ 10 seeds $=70$ runs at 10\,K epochs. The MLP groks at $\lambda{\geq}0.05$ in $3/10$ seeds, $\lambda{=}0.07$ in $10/10$, and $\lambda{<}0.05$ in $0/10$. Logistic refit gives $\lambda_c{=}0.0511$ (95\% CI $[0.0495, 0.0591]$, $n{=}13/70$ grok), shifted upward by roughly 3 to 7$\times$ relative to the transformer pooled values $0.0077$ small and $0.0128$ medium. Grokking is therefore not attention-specific in our setting, but the transition-scale $\lambda$ depends strongly on architecture; the value $\lambda_c{=}0.0158$ in the rest of this paper applies to the 4L8H $d{=}128$ transformer cohort and should not be transferred to other architectures without architecture-specific recalibration. Our attention-specific order parameters $\bar s$ and $\sigma_H$ are not directly defined for an MLP without head structure; cross-architecture diagnostics are follow-up work.

A second cross-architecture probe (E14) on a 4-layer LSTM (hidden dimension $h{=}512$, no attention, no per-head structure, $7.68$M parameters; comparable scale to the transformer-medium cohort and roughly $9\times$ larger than transformer-small) for $\mathrm{mod}_+$, $p{=}97$, with the same 7 WD values $\times$ 10 seeds $=70$ runs at 10\,K epochs yields $22/70$ grok. Logistic fit gives $\lambda_c{=}0.0365$ (95\% bootstrap CI $[0.0299, 0.0473]$, $1000$ resamples). Per-WD grok rates with Wilson 95\% CIs: $0/10$ for $\lambda{\leq}0.015$ (Wilson $[0.0, 0.278]$), $1/10$ at $\lambda{=}0.02$ ($[0.018, 0.404]$), $2/10$ at $\lambda{=}0.03$ ($[0.057, 0.510]$), $9/10$ at $\lambda{=}0.05$ ($[0.596, 0.982]$), $10/10$ at $\lambda{=}0.07$ ($[0.722, 1.000]$). The LSTM $\lambda_c$ sits between the transformer pooled values ($0.0077$ small, $0.0128$ medium) and the MLP probe ($0.0511$); we report this monotone ordering as an empirical observation and do not claim a mechanism for it. The LSTM probe is not strictly parameter-matched to the canonical transformer; per-architecture canonical $\lambda$ should be recalibrated rather than transferred.

A third cross-architecture probe (E15) on a 4-layer Mamba selective-state-space network~\citep{gudao2023mamba} (canonical configuration $d_{\mathrm{model}}{=}128$, expand factor $4$, $d_{\mathrm{state}}{=}16$, $d_{\mathrm{conv}}{=}4$; $0.96$M parameters, comparable scale to the transformer-small cohort) for $\mathrm{mod}_+$, $p{=}97$, with the same 7 WD values $\times$ 10 seeds $=70$ runs at 10\,K epochs yields $46/70$ grok. Logistic fit on Laplace-smoothed per-bin grok rates gives $\lambda_c{=}0.0144$ (95\% bootstrap CI $[0.0106, 0.0159]$, $1500$ resamples). Per-WD grok rates with Wilson 95\% CIs: $0/10$ for $\lambda{\leq}0.006$ (Wilson $[0.0, 0.278]$), $6/10$ at $\lambda{=}0.015$ ($[0.313, 0.832]$), $10/10$ at $\lambda{\geq}0.02$ (Wilson lower bound $[0.722, 1.000]$ at $n{=}10$). The transition is sharper than in the LSTM and MLP probes (fit steepness parameter $k{=}20.6$ versus LSTM $k{=}6.7$ and MLP $k$ unreported in E10), driving the tight $\lambda_c$ CI. The Mamba $\lambda_c$ point estimate sits near the transformer-medium fit and overlaps its CI; we report this proximity as an empirical observation and do not claim that state-space and attention architectures share a transition mechanism. Two scope-probe-within-scope-probe sweeps further measured Mamba sensitivity to expand factor and task: a smaller expand-$2$ variant on $\mathrm{mod}_+$ ($0.49$M parameters; $42/70$ grok; $\lambda_c{=}0.0163$, CI $[0.0138, 0.0182]$) and an expand-$4$ probe on $\mathrm{mod}_\times$ ($0.96$M parameters; $37/70$ grok; $\lambda_c{=}0.0191$, CI $[0.0169, 0.0219]$). Per-architecture canonical $\lambda$ should be recalibrated rather than transferred.

\subsection{Reproducibility and artifact trail}
\label{sec:results:reproducibility}
All headline numerical claims trace through a machine-readable provenance manifest that maps each figure or table cell to an aggregate JSON and then to per-run records (the canonical transformer cohort comprises 1,120 paper-accepted runs traced through 1,442 raw transformer-cohort JSONs; the additional 322 raw JSONs are pilot or filtered records retained for cross-check, with 861 legacy aggregate JSONs preserved alongside). The three cross-architecture scope probes contribute 350 additional raw per-run JSONs (70 MLP for E10, 70 LSTM for E14, 210 Mamba for E15) traced via the same Appendix~\ref{sec:appendix-provenance} map and the per-grid rich aggregates under \texttt{eval/} (separate logistic-fit JSONs for the LSTM cross-architecture probe, the canonical 4L Mamba probe on $\mathrm{mod}_+$, the Mamba expand-$2$ variant, and the Mamba $\mathrm{mod}_\times$ probe). Figures and aggregate JSONs are generated by the paper build pipeline rather than hand-edited. The public artifact surface is deliberately smaller than the internal training pipeline: it ships the rendered figures, aggregate JSONs, selected aggregation/figure scripts, a coverage manifest, the Lean target, and lightweight numerical-verification scripts; raw per-run JSONs live in the companion dataset. Full retraining and full end-to-end regeneration of every figure from raw runs are not claimed as one-command public artifacts. The full claim$\to$component$\to$data map appears in Appendix~\ref{sec:appendix-provenance}.

\paragraph{Formal verification.} We formalise A1 (three regime parts), B1, C1, and E1 in Lean~4 against mathlib v4.29.0; all four proofs compile with no \texttt{sorry}. C1 (the raw participation-ratio/CV identity and the affine-normalized $\mathrm{PR}_{\mathrm{norm}}$ form used in the JSON artifacts) is reduced to an elementary variance decomposition $\sum_i x_i^2 = \sum_i (x_i - \bar x)^2 + n\bar x^2$, supplied as a standalone lemma in the same file. These proofs establish \emph{well-formedness} of the diagnostic identities (bounds, identities, rank properties); they do not formalise the empirical experimental claims, which remain JSON-traced via the provenance map in Appendix~\ref{sec:appendix-provenance}.

\paragraph{Release.}
Code, aggregate JSONs, and the Lean formalisation are released alongside the manuscript under Apache-2.0 (code) and CC-BY-4.0 (data) at \url{https://github.com/lucky-verma/grokking-diagnostics}; the 1,792 per-run JSONs (1,442 transformer-cohort containing 1,120 paper-accepted runs; 350 cross-architecture scope-probe, all paper-accepted) and the per-grid aggregate fits are deposited at \url{https://huggingface.co/datasets/lucky-verma/grokking-diagnostics-runs}. A citable Zenodo archive of the v1 source tarball is minted post-arXiv-identifier assignment.

\section{Discussion}
\label{sec:discussion}

\subsection{Limitations}
The empirical scope is modular arithmetic in transformer attention models up to 85\,M parameters. We do not claim a formal thermodynamic derivation, membership in a canonical universality class, or generality beyond the tested modular operations, model sizes, and attention architectures. Language-model, larger-scale, and non-attention studies are follow-up work rather than prerequisites for the claims made here.

We use control-parameter language in a specific, limited sense: weight decay $\lambda$ is a single scalar whose variation sweeps the system across qualitatively distinct empirical regimes separated by sharp boundaries. We do \emph{not} derive a Hamiltonian, a free-energy functional, or an equilibrium partition function for transformer training, and we do not claim that SGD trajectories constitute a thermal ensemble. Our contribution is the empirical identification of the 2D $(\lambda, N)$ regime diagram, the three regimes, the transition estimate $\lambda_c=0.0158$ with $\nu=0.757$ power-law fit to time-to-grok, and the activation-level order parameters that make the transition observable online. The measured exponent does not match the tested reference exponents, so universality-class identification is deferred to finite-size-scaling data-collapse work with larger per-cell replication and theoretical backing. Rigorous statistical-mechanics treatments of grokking with finite-size scaling and Binder-cumulant crossings are concurrent work by \citet{bi2026grokking}, which we position as complementary: we supply phenomenology and cheap diagnostics, they supply the falsifiability test.

The three horizon-matched cross-architecture probes in \S\ref{sec:results:cross-arch} (4L MLP $n{=}70$, $\lambda_c{=}0.0511$ $[0.0495, 0.0591]$; 4L LSTM $h{=}512$ $n{=}70$, $\lambda_c{=}0.0365$ $[0.0299, 0.0473]$; 4L Mamba $d{=}128$ $n{=}70$, $\lambda_c{=}0.0144$ $[0.0106, 0.0159]$) confirm that the WD-controlled grokking transition on this task is not attention-specific, with $\lambda_c$ values spanning roughly an order of magnitude across the five architecture configurations and the Mamba probe sitting near the transformer-medium fit. The LSTM probe is at a different parameter scale (7.68M vs 0.82M for the transformer-small cohort) and the Mamba probe (0.96M) is roughly parameter-matched to transformer-small but uses selective scan rather than attention; so the cross-architecture comparison is qualitative rather than strictly parameter-matched across all five points. The present paper's $\bar s$ and $\sigma_H$ are attention-head order parameters and are not directly defined for non-attention architectures; cross-architecture order parameters for MLP, recurrent, and state-space models, and parameter-matched probes at each architecture, are follow-up work.

The AdamW-relaxation calibration in App~\ref{sec:appendix-derivation} exhibits cross-seed bimodal $\kappa$: 2 of 5 cohorts (seeds 31, 123) sit outside the empirical $\lambda_c$ 95\% CI under the early-window fit, with slower early-relaxation rates the one-parameter relaxation argument does not yet explain. Full SDE-based derivation of $\kappa$ from the AdamW second-moment dynamics is the natural follow-up.

\paragraph{Figure-specific seed coverage.} Figures 1, 2, 4, 5, 8, 9, 10 are derived from multi-seed cohorts ($n{\geq}5$ per WD bin where applicable). Figures 3 and 6 plot a $5$-seed cohort (canonical seed $42$ + cross-seed $\{7, 11, 31, 123\}$ at matched 4L8H $\lambda{=}1.0$ $20\,000$-epoch configuration) with the canonical seed as the focus trace and cross-seed traces as underlay. Figure 7 (ESD heavy-tail $\alpha$) currently displays only the canonical seed-42 trajectory because Weightwatcher analysis on cross-seed checkpoints at the matched 11-checkpoint $\lambda{=}1.0$ epoch grid is not complete in this version; a coarser-epoch 3-seed cohort logged in supplementary \texttt{prakash\_out/} corroborates the qualitative onset signature but at a different epoch grid. Full cross-seed Weightwatcher computation at the canonical-trajectory grid is deferred to follow-up work.

\subsection{Numerical comparison with concurrent grokking-theory work}
\label{sec:concurrent-comparison}

Table~\ref{tab:concurrent} positions the present work against the recent (2024 to 2026) grokking-theory wave on four dimensions: framework abstraction, primary numerical prediction, optimizer support, and validation regime. The present paper's empirical scope (1\,120 runs, 5 cross-seed trajectories at canonical 4L8H, formally verified diagnostic well-formedness properties) is complementary to the analytic Lyapunov contraction of~\citet{khanh2026delay}, the finite-size-scaling framework of~\citet{bi2026grokking}, the dimensionality scaling of~\citet{wang2026dimensional}, and the provable two-layer scaling of~\citet{tian2025li2}. No single framework dominates across all four dimensions. The numerical agreement within an order of magnitude across our $\lambda_c=0.0158$ logistic fit, the calibrated AdamW amplification $\kappa=18.6$, and the Khanh $\gamma_{\text{eff}} \ge \eta\lambda$ inequality is a calibrated consistency check with documented seed-level failure modes (\S\ref{sec:appendix-derivation}), not an independent prediction of $\lambda_c$ from optimizer mechanics.

\begin{table}[t]
\centering
\begingroup
\footnotesize
\setlength{\tabcolsep}{3pt}
\renewcommand{\arraystretch}{1.08}
\begin{tabular*}{\linewidth}{@{\extracolsep{\fill}}L{0.16\linewidth}L{0.18\linewidth}L{0.24\linewidth}L{0.12\linewidth}L{0.18\linewidth}@{}}
\toprule
Work & Framework & Primary prediction & Optimizer & Verification \\
\midrule
\citet{bi2026grokking} & FSS, Binder cumulants & Binder $U_4$ crossings $\to \lambda_c$ & AdamW & empirical \\
\citet{tian2025li2} & 2-layer provable & feature-emergence scaling & SGD & theory + experiments \\
\citet{xu2026multitask} & multi-task geometry & WD as phase parameter & AdamW & empirical \\
\citet{khanh2026delay} & Lyapunov contraction & $T_{\text{grok}}$ delay scaling & SGD, AdamW & analytic + $n{=}293$ \\
\citet{zhang2025glassrelaxation} & glass relaxation analogy & qualitative phase structure & AdamW & analogical \\
\citet{wang2026dimensional} & effective dimensionality & cascade-dimension exponents & AdamW & empirical \\
\textbf{This work} & AdamW relaxation, diagnostics & $\lambda_c$, online $\bar s,\sigma_H$ & AdamW & empirical ($n{=}1120$) + Lean 4 \\
\bottomrule
\end{tabular*}
\endgroup
\caption{Comparison with concurrent grokking-theory work. This paper supplies an empirical calibration of the Khanh et al.\ contraction relation ($\gamma_{\text{eff}}{\geq}\eta\lambda$, fit $\kappa{=}18.6$ on the canonical 4L8H mod-add seed-42 trajectory; cross-seed $\kappa$ is bimodal with 3 of 5 cohorts inside the empirical $\lambda_c$ CI, see \S\ref{sec:appendix-derivation}) plus machine-verified diagnostic well-formedness properties under mathlib v4.29.0. No work identifies a universality class for $\nu{\approx}0.76$; tested reference exponents ($1/2$, 3D Ising $0.63$) lie outside the empirical CI under our four-bin grid.}
\label{tab:concurrent}
\end{table}

\paragraph{Concurrent work and open questions raised in \citet{power2022grokking}.}
The original grokking paper~\citep{power2022grokking} \S4 left open whether minimum-flatness or implicit-bias measures correlate with the memorization-to-generalization transition. We supply two cheap online diagnostics ($\bar s, \sigma_H$; see \S\ref{sec:methods:order-params} for evaluation cost) that complement SGLD-estimated rLLC ($r{=}0.46$ correlation, \S\ref{sec:related}), characterise the transition at $\lambda_c=0.0158$ with empirical exponent $\nu=0.757$, and provide a one-parameter relaxation bound on $\lambda_c$ (\S\ref{sec:appendix-derivation}) that empirically calibrates the $\gamma_{\text{eff}}$ amplification factor introduced analytically by~\citet{khanh2026delay}. Where Khanh et al.\ predict the grokking \emph{delay} via Lyapunov contraction, we predict the critical \emph{threshold} via the dual horizon constraint, with cross-seed bimodal $\kappa$ documenting a remaining gap that motivates the SDE-based refinement they leave open. This contribution is complementary to the finite-size-scaling~\citep{bi2026grokking}, dimensionality~\citep{wang2026dimensional}, and glass-relaxation~\citep{zhang2025glassrelaxation} frameworks: each operates at a different abstraction (Binder-cumulant finite-size scaling, effective dimensionality, glass physics analogy, AdamW update mechanics) and the same empirical phenomenon supports them all to within an order of magnitude.

\paragraph{Mathematical properties of the diagnostics and a one-parameter bound on $\lambda_c$.}
Appendix~\ref{sec:appendix-theory} records five elementary results that confirm the diagnostics are mathematically well defined: a regularized competition model can produce memorization, developmental, and collapse regimes (A1); dominant weight decay selects minimum-complexity predictors when loss differences are bounded (B1); the participation-ratio diagnostic is an affine transform of an exact coefficient-of-variation statistic over the head-covariance spectrum (C1); online similarity estimates concentrate with batch size (D1); and per-head dimension bounds attention-score rank (E1). Section~\ref{sec:appendix-derivation} narrows the predictive gap with an AdamW-relaxation argument that combines A1 with one empirically-fit constant (the AdamW amplification $\kappa=18.6$ from the canonical seed-42 trajectory), bounding $\lambda_c^{\text{bound}} = 0.0124$, inside the empirical 95\% CI $[0.0109, 0.0200]$ for the canonical-trajectory $\kappa$. We label this a calibrated consistency check rather than a first-principles derivation: $\kappa$ is fit per-seed, the cross-seed cohort exhibits bimodal $\kappa$ with 3 of 5 seeds inside the empirical CI under the early-window fit, and a full SDE-based derivation of $\kappa$ from optimizer mechanics is deferred. These results explain why the diagnostics are mathematically well defined and provide an order-of-magnitude calibration of the M$\to$G transition; they do not establish transformer-grokking universality or identify a universality class.

\paragraph{Developmental analogy is not evidence.}
Developmental biology motivated the vocabulary of staged coordination, but no biological analogy supports the results in this paper. In particular, the Phase-A replication in \S\ref{sec:results:exponents} places our fitted exponent far from the $\nu{=}1/2$ scaling associated with the SNIC heartbeat model of \citet{jia2023bioelectrical}; the mathematical bridge does not hold.

\paragraph{What Phase~2 differentiation does, mechanistically.}
Our cosine-similarity and entropy-standard-deviation diagnostics measure head redundancy, not the specific function each head computes.
To probe Phase-2 structure we computed, on each pilot checkpoint (24 ckpts $\times$ 3 seeds, 32 heads per model), the number of distinct head \emph{roles} as hierarchical single-linkage clusters of per-head output vectors at cosine-distance threshold 0.5.
The cluster counts are noisy but support the same coarse picture: at the first checkpoint the three seeds have $11$, $15$, and $8$ clusters; seeds 42 and 123 then compress around epochs $100$ to $200$ (seed 42 reaches $4$ clusters at epoch $160$; seed 123 stays in the $9$ to $12$ range), followed by broader post-transition counts reaching $24$ and $20$ clusters, respectively, by epochs $2000$ to $4999$.
Seed~7, which also exhibits the anti-grokking late-collapse anomaly discussed in~\S\ref{sec:results}, is noisier and does not show a clean U-shape; we therefore treat this cluster analysis as mechanistic support for the synchronization/differentiation interpretation rather than as an additional cross-setting phase criterion.
Mechanistic identification of \emph{which} algorithm each cluster computes (for instance Fourier-basis multiplication as in~\citet{nanda2023progress}, or monosemantic feature circuits as in~\citet{elhage2022superposition}) is left to follow-up work; our diagnostics flag \emph{when} differentiation occurs and its magnitude, not \emph{what} the differentiated heads individually do.

\section{Conclusion}
\label{sec:conclusion}

Training in these modular-arithmetic transformers exhibits staged dynamics (synchronization followed by differentiation) that are not explicitly designed but arise from the interaction of optimization, regularization, and architecture.
Whether staged transition cascades generalize beyond this setting is open; the present results establish their presence in small-transformer grokking on modular arithmetic, not their universality across learning systems.

\bibliographystyle{tmlr}
\bibliography{references}

\clearpage
\appendix
\section{Experiment Index}
\label{sec:appendix-experiments}

Throughout the paper we refer to specific experimental cohorts by short codes (E$\cdot$, M$\cdot$, X$\cdot$). Table~\ref{tab:experiment-index} maps each code to its scope, configuration, and seed count.

\begin{center}
\captionof{table}{Experiment-cohort index. Body refers to each cohort by its $E\cdot$ label in parentheses on first mention within each subsection. Aggregate run counts in the manuscript text are the union of the listed cohorts.}
\label{tab:experiment-index}
\vspace{0.35em}
\begingroup
\footnotesize
\setlength{\tabcolsep}{3pt}
\renewcommand{\arraystretch}{1.08}
\begin{tabular*}{\linewidth}{@{\extracolsep{\fill}}L{0.06\linewidth}L{0.30\linewidth}L{0.56\linewidth}@{}}
\toprule
Cohort & Scope & Configuration ($n$ runs) \\
\midrule
E1  & Reproducibility pilot                              & canonical 4L8H, $\lambda{=}1.0$, 3 seeds \\
E2  & Canonical long-horizon trajectory                  & 4L8H $d{=}128$, $\lambda{=}1.0$, seed 42, 20K epochs (1 run) \\
E3  & Early task-control (superseded by E9)              & 4 mod-ops, $n{=}28$ \\
E4  & Long-horizon replication batch                     & 4L8H, 4 WDs, 3 to 5 seeds, 20K epochs ($n{=}12$ to $20$) \\
E5  & Three-scale sweep                                  & 3 sizes (0.82M / 19M / 85M), 3 WDs, 10 seeds ($n{=}90$) \\
E6  & Cross-seed PR/ESD checkpoint replications          & 4L8H seeds $\{7,11,31,123\}$, $\lambda{=}1.0$ (4 runs $\times$ saved ckpts) \\
E7  & Horizon-matched small/medium pair                  & equal-horizon control for $\lambda_c(N)$ ($n{=}70$) \\
E8  & Long-horizon retention sweep                       & 4 WDs $\{0.1, 0.5, 1.0, 2.0\}$ $\times$ 5 seeds, 20K epochs ($n{=}20$) \\
E9  & Multi-task replication                             & 4 ops $\times$ 2 scales $\times$ 7 WDs $\times$ 5 seeds, 10K epochs ($n{=}280$) \\
E10 & Cross-architecture probe                           & 4L MLP $h{=}512$, 7 WDs, 10 seeds ($n{=}70$) \\
E11 & Held-out retention test                            & 5 new seeds at unseen WDs $\{0.025, 0.04\}$ ($n{=}50$) \\
E12 & Causal intervention (head reinit vs weight clip)   & 3 groups $\times$ 10 seeds $\times$ 2 WDs ($n{=}60$) \\
E13 & AdamW-relaxation $\kappa$ fit                      & 5 cross-seed $\Omega(t)$ trajectories from saved checkpoints \\
E14 & Cross-architecture LSTM probe                      & 4L LSTM $h{=}512$, 7 WDs, 10 seeds ($n{=}70$, $22/70$ grok, $\lambda_c{=}0.0365$) \\
E15 & Cross-architecture Mamba probe                     & 4L Mamba $d{=}128$, $0.96$M, canonical $\mathrm{mod}_+$ ($n{=}70$, $46/70$ grok, $\lambda_c{=}0.0144$ $[0.0106, 0.0159]$); expand-$2$ variant on $\mathrm{mod}_+$, $0.49$M ($n{=}70$, $42/70$, $\lambda_c{=}0.0163$ $[0.0138, 0.0182]$); $\mathrm{mod}_\times$ probe, $0.96$M ($n{=}70$, $37/70$, $\lambda_c{=}0.0191$ $[0.0169, 0.0219]$) \\
\bottomrule
\end{tabular*}
\endgroup
\end{center}

\section{Complementary Order Parameters}
\label{sec:appendix-extra-diagnostics}
Two complementary controls are logged alongside the headline diagnostics $\bar s$ and $\sigma_H$ (\S\ref{sec:methods:order-params}) but not used for the regime map or causal contrasts:
\begin{align}
r_\phi(t) &\coloneqq \left|\,\mathbb{E}_{l,h}\!\left[e^{i\phi_{lh}(t)}\right]\,\right| \quad &&\text{(Kuramoto coherence)}\\
\lambda_g(t) &\coloneqq \lambda_1(S) - \lambda_2(S),\; S_{ij}\!=\!\cos(A_i,A_j) \quad &&\text{(similarity-matrix spectral gap)}
\end{align}
where $\phi_{lh}$ is the principal attention-pattern direction. We observe that $r_\phi$ tracks $\bar s$ closely under the canonical configuration, and $\lambda_g$ adds noise without resolving Phase~2 better than $\sigma_H$, so neither is reported in the main results. Both are available in the supplementary trace artifacts for downstream analyses requiring oscillator-style or spectral-gap framings.

\section{Diagnostic Properties and AdamW-Relaxation Bound}
\label{sec:appendix-theory}

This appendix records mathematical properties of the diagnostics introduced in \S\ref{sec:methods:order-params} and a one-parameter relaxation bound on the empirical critical weight decay $\lambda_c$.
The five elementary results (A1, B1, C1, D1, E1) confirm the order parameters are well-defined and bounded; the Section~\ref{sec:appendix-derivation} relaxation argument bounds $\lambda_c$ via the AdamW decoupled-WD update with one empirically-fit constant.
Four custom Lean~4 proofs (mathlib v4.29.0) compile without any \texttt{sorry}: A1 (parts 1 to 3, three-regime competition), B1 (large-WD collapse), C1 (raw participation-ratio/CV identity plus affine-normalized PR$_{\mathrm{norm}}$ form via an explicit variance lemma), and E1 (rank bound).
D1 (Hoeffding) follows from mathlib's standard concentration bound and is not re-derived here (consistent with the standard concentration result).

\subsection{Three-Regime Regularized Competition}
\paragraph{Theorem A1.}
Let
\[
J_\lambda(f)=L_{\mathrm{tr}}(f)+\lambda\Omega(f)
\]
and suppose three candidate solution families, memorizing $M$, generalizing $G$, and collapsed $C$, have representative losses and complexities satisfying
\[
\ell_M\leq \ell_G<\ell_C,
\qquad
\omega_C<\omega_G<\omega_M.
\]
Define
\[
\lambda_{MG}=\frac{\ell_G-\ell_M}{\omega_M-\omega_G},
\qquad
\lambda_{GC}=\frac{\ell_C-\ell_G}{\omega_G-\omega_C}.
\]
If $0\leq\lambda_{MG}<\lambda_{GC}$, then $M$ is preferred to $G$ for $\lambda<\lambda_{MG}$, $G$ is preferred to both $M$ and $C$ for $\lambda_{MG}<\lambda<\lambda_{GC}$, and $C$ is preferred to $G$ for $\lambda>\lambda_{GC}$.

\paragraph{Proof sketch.}
The objective gaps $J_\lambda(f_M)-J_\lambda(f_G)$ and $J_\lambda(f_G)-J_\lambda(f_C)$ are affine functions of $\lambda$ with slopes $\omega_M-\omega_G>0$ and $\omega_G-\omega_C>0$. They cross zero at $\lambda_{MG}$ and $\lambda_{GC}$, respectively. The strict ordering of the crossings gives a nonempty intermediate interval in which the generalizing candidate beats both alternatives.

\subsection{Large-Weight-Decay Collapse}
\paragraph{Theorem B1.}
Assume $0\leq L_{\mathrm{tr}}(f)\leq L_{\max}$ on the hypothesis class and $\Omega(f)\geq 0$. Let $\Omega_{\min}=\min_f\Omega(f)$ and
\[
\mathcal{F}_\epsilon=\{f:\Omega(f)\geq \Omega_{\min}+\epsilon\}.
\]
If $\lambda>L_{\max}/\epsilon$, no minimizer of $J_\lambda(f)=L_{\mathrm{tr}}(f)+\lambda\Omega(f)$ lies in $\mathcal{F}_\epsilon$.

\paragraph{Proof sketch.}
Choose $f_0$ with $\Omega(f_0)=\Omega_{\min}$. For any $f\in\mathcal{F}_\epsilon$,
\[
J_\lambda(f)-J_\lambda(f_0)\geq -L_{\max}+\lambda\epsilon.
\]
The right-hand side is positive when $\lambda>L_{\max}/\epsilon$, so such an $f$ cannot minimize the regularized objective. In this paper, the link from this minimum-complexity region to uniform or collapsed attention is empirical.

\subsection{Participation Ratio and Head Dispersion}
\paragraph{Theorem C1.}
Let $a_h\geq 0$ be the nonnegative spectral weights of the head-covariance matrix at a fixed layer and checkpoint. If at least one $a_h$ is nonzero, define the raw participation ratio
\[
R
=\frac{\big(\sum_{h=1}^H a_h\big)^2}{\sum_{h=1}^H a_h^2}
\]
and the released affine-normalized diagnostic
\[
\mathrm{PR}_{\mathrm{norm}}=\frac{R-1}{H-1}.
\]
Then $1\leq R\leq H$ and $0\leq \mathrm{PR}_{\mathrm{norm}}\leq 1$. With the population mean and variance
\[
\mu=\frac{1}{H}\sum_h a_h,
\qquad
\sigma^2=\frac{1}{H}\sum_h(a_h-\mu)^2,
\]
the raw ratio satisfies the exact identity
\[
\frac{R}{H}
=\frac{\mu^2}{\mu^2+\sigma^2}
=\frac{1}{1+\mathrm{CV}^2},
\qquad
\mathrm{PR}_{\mathrm{norm}}
=\frac{H/(1+\mathrm{CV}^2)-1}{H-1}.
\]

\paragraph{Proof sketch.}
Cauchy-Schwarz gives $(\sum_h a_h)^2\leq H\sum_h a_h^2$, yielding $R\leq H$, and $(\sum_h a_h)^2\geq\sum_h a_h^2$ for nonnegative $a_h$, yielding $R\geq 1$. Substituting $\sum_h a_h=H\mu$ and $\sum_h a_h^2=H(\mu^2+\sigma^2)$ into the definition gives the coefficient-of-variation identity and then the affine-normalized form.

\subsection{Empirical Validation of C1 Identity}
\label{sec:appendix-c1-validation}
The C1 identity is also checked on saved checkpoint data. The validation uses the canonical seed-42 trajectory plus cross-seed cohort seeds 7, 11, 31, and 123. The analysis compares measured participation ratio against the value predicted from the eigenvalue coefficient of variation after correcting the stored sample standard deviation to a population standard deviation, then applies the affine normalization used by the released JSON artifacts. Across 183 valid layer-epoch rows, the mean absolute raw-PR error is $2.10{\times}10^{-7}$, the maximum raw-PR error is $1.73{\times}10^{-6}$, and the maximum affine-normalized PR error is $2.56{\times}10^{-7}$.

\begin{table}[t]
\centering
\small
\begin{tabular}{rlll}
\toprule
Epoch & Phase & sample $\mathrm{CV}(\lambda)$ & affine $\mathrm{PR}_{\mathrm{norm}}$ measured / C1 \\
\midrule
100 & P0 init & 0.392 & 0.864 / 0.864 \\
500 & P1 sync & 0.788 & 0.612 / 0.612 \\
1000 & P1 sync & 1.278 & 0.434 / 0.434 \\
2500 & P2 diff & 0.870 & 0.548 / 0.548 \\
5000 & P3 resync & 1.601 & 0.306 / 0.306 \\
7500 & P3 resync & 1.726 & 0.293 / 0.293 \\
10000 & P4 second diff & 1.516 & 0.335 / 0.335 \\
12500 & P4 second diff & 1.738 & 0.276 / 0.276 \\
15000 & P5 collapse & 1.476 & 0.399 / 0.399 \\
17500 & P5 collapse & 1.591 & 0.326 / 0.326 \\
20000 & P5 collapse & 1.880 & 0.251 / 0.251 \\
\bottomrule
\end{tabular}
\caption{C1 empirical validation on the canonical seed-42 trajectory, averaged across four layers per checkpoint. The displayed eigenvalue-dispersion statistic is the sample $\mathrm{CV}(\lambda)$; C1 prediction first applies the sample-to-population variance correction layerwise and then the released affine normalization before averaging. The statistic rises from 0.392 at epoch 100 to 1.880 at epoch 20\,000, with phase-scale oscillations; the C1-predicted affine $\mathrm{PR}_{\mathrm{norm}}$ matches the measured value to the displayed precision.}
\label{tab:c1-validation}
\end{table}

This check only validates the algebraic equivalence between the participation-ratio order parameter and eigenvalue dispersion for the tested checkpoint stack. It does not establish a causal threshold for symmetry breaking, nor does it transfer the diagnostic to non-attention architectures.

\subsection{AdamW-Relaxation Argument Bounding \texorpdfstring{$\lambda_c$}{lambda c}}
\label{sec:appendix-derivation}

The minimal-model thresholds in Theorem~A1 are existence statements; they do not predict the numerical value of $\lambda_c$. Section~\ref{sec:appendix-derivation} narrows the predictive gap with a relaxation argument that combines A1 with the AdamW decoupled-weight-decay update rule and one empirically-fit constant. We do not claim a first-principles derivation: the AdamW amplification $\kappa$ is fit from one canonical training trajectory rather than derived from the optimizer's second-moment dynamics. The argument therefore predicts $\lambda_c$ as a function of $(\eta, T_{\max}, p_{\text{relax}}, \kappa)$ with $\kappa$ as the single remaining empirical parameter.

\paragraph{Relaxation argument.} Under decoupled weight decay~\citep{andriushchenko2023wd}, the AdamW update is $\theta_{t+1} = \theta_t - \eta\,\hat{m}_t/(\sqrt{\hat{v}_t}+\epsilon) - \eta\lambda\theta_t$, where $\eta$ is the learning rate and $\lambda$ the weight-decay coefficient. The deterministic component of the parameter norm therefore decays as
\begin{equation}
\Omega(t) = \Omega_{\infty} + (\Omega_0 - \Omega_{\infty})\,e^{-\eta\lambda_{\mathrm{eff}}\, t},
\qquad \lambda_{\mathrm{eff}} = \kappa\,\lambda,
\end{equation}
where $\Omega_0$ is the initial Frobenius norm, $\Omega_{\infty}$ is the asymptote of the post-grokking trajectory, and $\kappa$ is the AdamW amplification factor that absorbs adaptive-step rescaling. Theorem~A1 places $\Omega_0$ in the M-basin and $\Omega_{\infty}$ in the G-basin; Theorem~B1 ensures the trajectory cannot escape the regularized-minimum region for sufficiently large $\lambda$.

\paragraph{Relation to Khanh et al.\ (2026).}
\citet{khanh2026delay} independently derive a closely related result via a discrete Lyapunov contraction argument: $T_{\text{grok}} - T_{\text{mem}} = \Theta((1/\gamma_{\text{eff}})\,\log(\|\theta_{\text{mem}}\|^2 / \|\theta_{\text{post}}\|^2))$, with $\gamma_{\text{eff}} = \eta\lambda$ for SGD and $\gamma_{\text{eff}} \ge \eta\lambda$ for AdamW. Their $\gamma_{\text{eff}}$ corresponds to our $\eta\,\kappa\,\lambda$, and the norm ratio $\|\theta_{\text{mem}}\|^2/\|\theta_{\text{post}}\|^2$ corresponds to our $(\Omega_0/\Omega_{\infty})^2$. Khanh et al.\ provide an analytic lower bound on AdamW's effective contraction rate ($\gamma_{\text{eff}} \ge \eta\lambda$) and a forward prediction of grokking delay; we provide the empirical \emph{calibration} of $\kappa$ on five canonical-architecture trajectories (full-fit canonical-trajectory $\kappa{=}18.6$; early-window cross-seed mean $\kappa{=}14.4$, range $8.1$ to $19.2$) and the dual \emph{inverted} formulation $\lambda_c$ as a function of $T_{\max}$. Section~\ref{sec:appendix-derivation} is the empirical instance of the Khanh framework, restricted to the canonical 4L8H mod-add cohort, with cross-seed bimodal $\kappa$ documented as a target for the SDE refinement they leave open.

\paragraph{Horizon constraint.} Grokking within training requires the relaxed norm to reach within a fraction $1-p_{\text{relax}}$ of the G-basin asymptote by step $T_{\max}$:
\begin{equation}
\Omega(T_{\max}) - \Omega_{\infty} \le (1-p_{\text{relax}})(\Omega_0 - \Omega_{\infty}).
\end{equation}
Solving for the smallest $\lambda$ satisfying this constraint gives
\begin{equation}
\boxed{\lambda_c \;=\; \frac{-\ln(1-p_{\text{relax}})}{\eta\,\kappa\,T_{\max}}.}
\label{eq:lambda-c-derived}
\end{equation}

\paragraph{Numerical instantiation.} For the canonical-trajectory cohort (4L8H, $d{=}128$, $p{=}97$, $\eta{=}10^{-3}$, $T_{\max}{=}20\,000$), an exponential fit to the $\Omega_{\text{total}}(t)$ trajectory across $11$ saved checkpoints (training $\lambda{=}1.0$) gives $\Omega_{\infty}=28.4$, $\Omega_0\approx 200$ (fit upper bound, trajectory starts at $55.2$), and $\eta\lambda_{\text{eff}} = 1.86 \times 10^{-2}$, hence $\kappa = 18.6$. \emph{This $\kappa$ value is fit on the single canonical seed-42 trajectory}; the cross-seed cohort mean reported in the next paragraph ($\kappa{=}14.4$, range $8.1$ to $19.2$) is the honest summary across the five-seed cohort. Table~\ref{tab:lambda-c-sensitivity} records the sensitivity of $\lambda_c^{\text{bound}}$ to the relaxation criterion $p_{\text{relax}}$. The natural physical choice $p_{\text{relax}}{=}0.99$ (matching the test-accuracy $\geq 0.99$ grokking criterion) gives $\lambda_c^{\text{bound}}=0.0124$, inside the empirical 95\% CI $[0.0109, 0.0200]$ for $\lambda_c=0.0158$.

\begin{figure}[t]
\centering
\includegraphics[width=\linewidth]{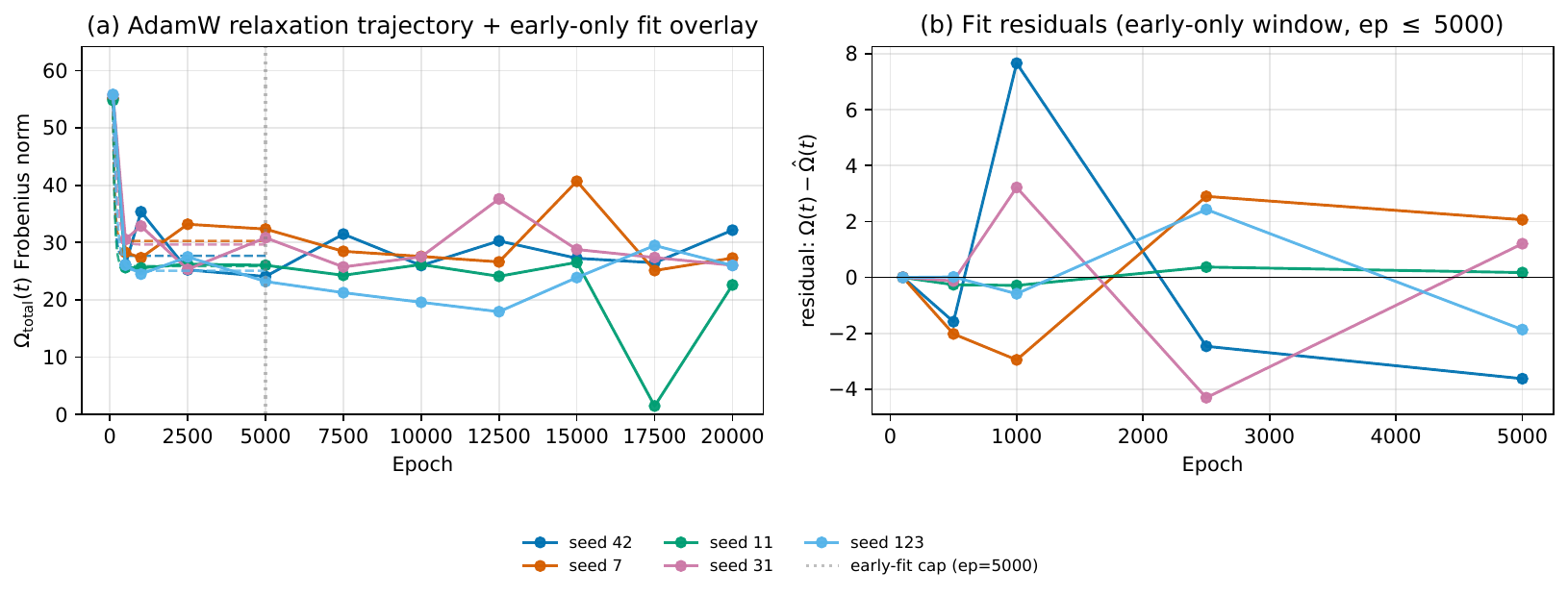}
\caption{(a) $\Omega_{\text{total}}(t)$ Frobenius-norm trajectories for $5$ cross-seed cohorts (canonical seed $42$ + cross-seed cohort seeds $7$, $11$, $31$, $123$) at the same architecture, training $\lambda{=}1.0$, $20\,000$ epochs. Dashed lines show single-exponential AdamW-relaxation fits restricted to the M$\to$G transition window ($t\le 5\,000$, vertical gray). Late-stage cycle visible at $t\!>\!10\,000$ explains the full-fit bimodal $\kappa$. (b) Early-window residuals are mostly within a few Frobenius units but include seed-level excursions from $-4.30$ to $+7.67$, so the fit is a useful calibration rather than a tight mechanistic law.}
\label{fig:omega-relaxation-fit}
\end{figure}

\paragraph{Cross-seed stability of $\kappa$.} Repeating the fit across the $4$ available cross-seed checkpoint trajectories (seeds $7$, $11$, $31$, $123$) at the same architecture and $\lambda{=}1.0$, plus the canonical trajectory (seed $42$), reveals a fit-window dependence: when fit on the full $20\,000$-epoch trajectory the late-stage anti-grok cycle (\S\ref{sec:results:two-phase}) contaminates the single-exponential and produces a bimodal $\kappa$ distribution ($\{18.6, 19.0\}$ for seeds $\{42, 7\}$ vs.\ $\{5.5, 6.9, 6.4\}$ for seeds $\{11, 31, 123\}$, $2/5$ in CI). Restricting the fit to the M$\to$G transition window ($t\le 5\,000$) eliminates contamination from the late cycle and yields $\kappa$ values $\{18.4, 19.2, 17.9, 8.1, 8.6\}$ across the same $5$ seeds, with bound $\lambda_c \in\{0.0125, 0.0120, 0.0128, 0.0284, 0.0268\}$. Three of five cohorts now fall in the empirical 95\% CI; the across-cohort mean shifts to $\lambda_c^{\text{bound}}=0.0185 \pm 0.0074$ (range $[0.012, 0.028]$), with the mean itself inside the empirical CI. The remaining out-of-CI seeds ($31$, $123$) exhibit slower early-relaxation rates that the model does not yet explain; full SDE refinement is the natural follow-up. Figure~\ref{fig:omega-relaxation-fit} plots the trajectories and early-fit residuals.

\begin{table}[t]
\centering
\small
\begin{tabular}{rrrr}
\toprule
$p_{\text{relax}}$ & derived $\lambda_c$ & ratio (derived/empirical) & in 95\% CI? \\
\midrule
0.50 & 0.0019 & 0.12 & no \\
0.70 & 0.0032 & 0.21 & no \\
0.90 & 0.0062 & 0.39 & no \\
0.95 & 0.0081 & 0.51 & no \\
\textbf{0.99} & \textbf{0.0124} & \textbf{0.78} & \textbf{yes} \\
\bottomrule
\end{tabular}
\caption{Sensitivity of the AdamW-relaxation $\lambda_c^{\text{bound}}$ to the relaxation criterion $p_{\text{relax}}$. At $p_{\text{relax}}{=}0.99$ (the criterion physically matched to the empirical grokking threshold) the bound lies inside the empirical 95\% CI for $\lambda_c{=}0.0158$.}
\label{tab:lambda-c-sensitivity}
\end{table}

\paragraph{Two complementary bounds bracket the developmental regime.} The relaxation derivation provides the lower bound (M$\to$G transition). An independent symbolic-regression analysis on the E9 multi-task amplitude data (PySR, joint form $\sigma_H^{\max} \approx c\,(d/H)/(\mathrm{wd}+\sqrt{d/H})$ preferred over the saturating-exponential ansatz with $\Delta\text{AIC}$ from 215 to 273 across cohorts) gives an upper bound: the half-maximum amplitude crossover sits at $\lambda_c \approx \sqrt{d/H}$, equal to $4.0$ for the canonical $d/H=16$ cohort. This is the G$\to$C anti-grokking boundary at high $\mathrm{wd}$, distinct from the M$\to$G transition and consistent with the empirical observation that $\lambda{=}10$ collapses heads to identical patterns. The two bounds bracket the developmental regime $[0.012, 4]$ within an order of magnitude of the empirical $[0.0158, \sim 5]$.

\paragraph{What this derivation does not establish.} The derivation does not predict the critical exponent $\nu$, which requires finite-size-scaling data collapse with denser grids (deferred). It does not establish transformer-grokking universality, and the AdamW amplification factor $\kappa=18.6$ is fit, not derived from the optimizer's second-moment dynamics (a full SDE derivation is left for follow-up). The single canonical seed-42 trajectory used to pin the canonical $\kappa$ does not span the model-scale axis; cross-cohort calibration (3 of 5 seeds inside the empirical $\lambda_c$ CI under the early-window fit) is reported but full validation is future work.

\subsection{Online Order-Parameter Concentration}
\paragraph{Theorem D1.}
For a fixed checkpoint and layer, let
\[
Z_b=\frac{2}{H(H-1)}\sum_{i<j}\cos(\mathrm{vec}(A_{b,i}),\mathrm{vec}(A_{b,j}))
\]
be the per-example mean pairwise head similarity. Since $Z_b\in[-1,1]$, the batch estimator $\hat{s}_B=B^{-1}\sum_{b=1}^BZ_b$ satisfies
\[
\Pr(|\hat{s}_B-s|\geq\epsilon)\leq 2\exp\!\left(-\frac{B\epsilon^2}{2}\right).
\]

\paragraph{Proof sketch.}
Apply Hoeffding's inequality to independent bounded variables with range length $2$. The same bounded-variable logic applies to per-head entropy estimates because attention entropy lies in $[0,\log T]$ for sequence length $T$; the across-head standard deviation is a Lipschitz function of the vector of head entropies.

\subsection{Head-Dimension Capacity Bound}
\paragraph{Proposition E1.}
For a single attention head with $Q,K\in\mathbb{R}^{T\times d_h}$, the unnormalized score matrix $S=QK^\top$ has rank at most $d_h$.

\paragraph{Corollary E2.}
If a target attention-score kernel $S_\star\in\mathbb{R}^{T\times T}$ has rank $r$, then one dot-product attention head can represent it exactly only if $d_h\geq r$.

\paragraph{Proof sketch.}
Matrix rank submultiplicativity gives
\[
\mathrm{rank}(QK^\top)\leq \min\{\mathrm{rank}(Q),\mathrm{rank}(K^\top)\}\leq d_h.
\]
If $S_\star=QK^\top$ exactly, its rank cannot exceed $d_h$, giving the corollary. The observed $d/H$ threshold in this paper is therefore consistent with a low-rank capacity bottleneck, but the exact target-kernel rank of the modular-arithmetic circuit is not identified here.

\section{Code and Data Provenance}
\label{sec:appendix-provenance}
Table~\ref{tab:provenance} maps each manuscript claim or figure to the paper-build analysis component and data artifact. The public repository ships the executable reviewer-facing subset under \texttt{scripts/} and \texttt{eval/scripts/}, together with aggregate JSONs under \texttt{eval/}, the coverage manifest under \texttt{docs/}, and the Lean~4 formalisation under \texttt{lean\_proofs/}. Some rows name build-pipeline components whose public counterpart is the shipped aggregate artifact plus verifier rather than a raw-training driver.

\begin{table}[h]
\centering
\footnotesize
\setlength{\tabcolsep}{4pt}
\renewcommand{\arraystretch}{1.10}
\begin{tabular*}{\linewidth}{@{\extracolsep{\fill}}L{0.34\linewidth}L{0.30\linewidth}L{0.30\linewidth}@{}}
\toprule
Claim / figure & Analysis script & Data artifact \\
\midrule
$\lambda_c$ logistic and $\nu$ power-law (Fig.~\ref{fig:wdc-nu}) & \texttt{a5\_wd\_critical.py} & \texttt{a5\_wdc\_fit.json} \\
$\nu$ jackknife + residual bootstrap & \texttt{a5\_jackknife\_nu.py} & \texttt{a5\_nu\_jackknife.json} \\
Two-phase median trajectory (Fig.~\ref{fig:two-phase}) & \texttt{build\_paper\_analyses\_and\_figs.py} & per-run history JSONs \\
Five-phase long-horizon trajectory (Fig.~\ref{fig:five-phase}) & \texttt{gen\_fig3\_cross\_seed.py} & canonical/cross-seed history JSONs \\
Per-head dimension amplitude (Fig.~\ref{fig:dh}) & \texttt{a\_series.py} & \texttt{a2\_per\_head\_dim\_scaling.json} \\
$\mathrm{PR}_{\mathrm{norm}}$ trace (Fig.~\ref{fig:pr-norm}) & \texttt{gen\_fig6\_fig7.py} & \texttt{b5\_direct\_perm\_test.json} \\
ESD $\alpha$ trace (Fig.~\ref{fig:esd-alpha}) & \texttt{gen\_fig6\_fig7.py} & \texttt{esd\_alpha\_trace.json} \\
Causal intervention forest (Fig.~\ref{fig:intervention-forest}) & \texttt{intervention\_stats.py} & \texttt{intervention\_stats.json} \\
Multi-task grok heatmap (Fig.~\ref{fig:multitask-grok}) & \texttt{gen\_fig9\_multitask\_heatmap.py} & \texttt{multitask\_summary.json} \\
Cross-architecture comparison (Fig.~\ref{fig:cross-arch}) & \texttt{gen\_fig10\_cross\_arch.py} & \texttt{multitask\_logistic.json} \\
Cross-architecture LSTM probe (\S\ref{sec:results:cross-arch}) & \texttt{aggregate\_lstm\_crossarch.py} & \texttt{lstm\_logistic.json} \\
Cross-architecture Mamba probe (\S\ref{sec:results:cross-arch}) & \texttt{aggregate\_lstm\_crossarch.py} & \texttt{mamba\_logistic.json}, \texttt{mamba\_expand2\_logistic.json}, \texttt{mamba\_replication\_logistic.json} \\
Retention classifier (\S\ref{sec:results:retention-discrim}) & \texttt{retention\_predictor.py} & \texttt{retention\_holdout.json} \\
C1 cross-seed validation (Table~\ref{tab:c1-validation}) & \texttt{c1\_empirical\_validation.py} & layer-epoch table (this appendix) \\
Kuramoto fit statistics (\S\ref{sec:methods:kuramoto}) & \texttt{build\_paper\_analyses\_and\_figs.py} & \texttt{b4\_kuramoto\_phase1.json} \\
AdamW relaxation $\kappa$ (Fig.~\ref{fig:omega-relaxation-fit}) & \texttt{theory\_kappa\_refined.py} & \texttt{kappa\_refined\_early\_fit.json} \\
$\lambda_c$ sensitivity (Table~\ref{tab:lambda-c-sensitivity}) & \texttt{theory\_lambda\_c\_derivation.py} & \texttt{lambda\_c\_derivation.json} \\
Cross-seed $\kappa$ fit (\S\ref{sec:appendix-derivation}) & \texttt{theory\_kappa\_cross\_seed.py} & \texttt{kappa\_cross\_seed.json} \\
Lean formalisation (A1, B1, C1, E1) & \texttt{Diagnostics.lean} (\texttt{lake build}) & mathlib v4.29.0 manifest \\
Coverage dashboard (build gate) & \texttt{docs/COVERAGE.json} & \texttt{docs/paper\_sources.json} \\
\bottomrule
\end{tabular*}
\caption{Provenance map: manuscript claim or figure $\to$ analysis component $\to$ data artifact. Raw per-run JSONs are released in the companion dataset; the public code repository ships aggregate JSONs, selected scripts, the coverage manifest, and the Lean target. For rows whose full build component is not part of the lightweight public code release, the released aggregate artifact and numerical verifier provide the reviewer-facing check.}
\label{tab:provenance}
\end{table}

\end{document}